\documentclass{article}

    \PassOptionsToPackage{numbers, compress}{natbib}

\usepackage[preprint]{neurips_2024}

\usepackage[utf8]{inputenc} 
\usepackage[T1]{fontenc}    
\usepackage{hyperref}       
\usepackage{url}            
\usepackage{booktabs}       
\usepackage{amsfonts}       
\usepackage{nicefrac}       
\usepackage{microtype}      
\usepackage{xcolor}         
\usepackage{amsmath, amsthm} 
\usepackage{subcaption, graphicx, multirow, arydshln, float}
\usepackage{sidecap} 
\usepackage[skip=5pt]{caption} 
\newtheorem{theorem}{Theorem} 

\newcommand{\norm}[1]{{\left\|{#1}\right\|}}
\def \bR {\mathbb{R}}

\title{Addressing Spectral Bias of Deep Neural Networks by Multi-Grade Deep Learning}

\author{%
  Ronglong Fang, Yuesheng Xu\thanks{Corresponding author: \texttt{y1xu@odu.edu}} \\
  Department of Mathematics and Statistics, Old Dominion University\\
  \texttt{\{rfang002, y1xu\}@odu.edu}
}

\begin{document}

\maketitle

\begin{abstract}
Deep neural networks (DNNs) have showcased their remarkable precision in approximating smooth functions. However, they suffer from the {\it spectral bias}, wherein DNNs typically exhibit a tendency to prioritize the learning of lower-frequency components of a function, struggling to effectively capture its high-frequency features. This paper is to address this issue. Notice that a function having only low frequency components may be well-represented by a shallow neural network (SNN), a network having only a few layers. By observing that composition of low frequency functions can effectively approximate a high-frequency function, we propose to learn a function containing high-frequency components by composing several SNNs, each of which learns certain low-frequency information from the given data. We implement the proposed idea by exploiting the multi-grade deep learning (MGDL) model, a recently introduced model that trains a DNN incrementally, grade by grade, a current grade learning from the residue of the previous grade only an SNN (with trainable parameters) composed with the SNNs (with fixed parameters) trained in the 
preceding grades as features.
We apply MGDL to synthetic, manifold, colored images, and MNIST datasets, all characterized by presence of high-frequency features. Our study reveals that MGDL excels at representing functions containing high-frequency information. Specifically, the neural networks learned in each grade adeptly capture some low-frequency information, allowing their compositions with SNNs learned in the previous grades effectively representing the high-frequency features. 
Our experimental results underscore the efficacy of MGDL in addressing the spectral bias inherent in DNNs. By leveraging MGDL, we offer insights into overcoming spectral bias limitation of DNNs, thereby enhancing the performance and applicability of deep learning models in tasks requiring the representation of high-frequency information. This study confirms that the proposed method offers a promising solution to address the spectral bias of DNNs. The code is available on GitHub: \href{https://github.com/Ronglong-Fang/AddressingSpectralBiasviaMGDL}{\texttt{Addressing Spectral Bias via MGDL}}.
\end{abstract}



\section{Introduction}

Deep neural networks (DNNs) have achieved tremendous success in various applications, including computer vision \cite{He2016}, natural language processing \cite{Vaswani2017}, speech recognition \cite{Nassif2019}, and finance \cite{Ozbayoglu2020}.
From a mathematical perspective, the success is mainly due to their high expressiveness, as evidenced by theoretical demonstrations showing their capability to approximate smooth functions with arbitrary precision  \cite{Arpit2017, Cybenko1989, Goodfellow2016, Hornik1989, zhang2021}. Various mathematical aspects of DNNs as an approximation tool were recently investigated in  \cite{daubechies2022nonlinear, huang2022error,li2024two, mhaskar2016deep, shen2022optimal, XuZhang2022, XuZhang2021, Xu_Zhang:IEEEIT2024, zhou2020universality}.
However, it was noted in \cite{Rahaman2019, Xu2019Ffrequencya} that the standard deep learning model, which will be called single grade deep learning (SGDL),  trains a DNN end to end leading to learning bias towards low-frequency functions. While this bias may explain the phenomenon where DNNs with a large number of parameters can achieve low generalization error \cite{Cao2019, XuZJ2018, Xu2022},
DNNs trained by SGDL struggle to capture high-frequency components within a function even though they can well-represent its low-frequency components. This bias may potentially limit the applicability of DNNs to problems involving high-frequency features, such as image reconstruction \cite{Fang2024Inexactell0, LuShenXu2006, Tang2021}, seismic wavefield modeling \cite{Wu2022SeismicWavefieldModeling},
high-frequency wave equations in homogenization periodic media \cite{Craster2010Highfrequencyhomogenization}, and high energy physics \cite{Perkins2000highenergyphysics}. Especially, in medical image reconstruction such as PET/SPECT, high-frequency components play a crucial role in determining image resolution, which is critical in clinical practice, as higher resolution leads to earlier and more accurate disease diagnosis. 


There have been some efforts to address this issue. 
A phrase shift DNN was proposed in \cite{Cai2020}, where the original dataset was first decomposed into subsets with specific frequency components, the high-frequency component was shifted downward to a low-frequency spectrum for learning, and finally, the learned function was converted back to the original high frequency. 
An adaptive activation function was proposed in \cite{Jagtap2020} to replace the traditional activation
by scaling it with a trainable parameter.
A multiscale DNN was introduced in \cite{cai2019, Liu2020}, in which the input variable was first scaled with different scales and then the multiscale variables were combined to learn a DNN. 
It was proposed in \cite{Tancik2020} first to map the input variable to Fourier features with different frequencies and then to train the mapped data by DNNs. 
All these approaches can mitigate the spectral bias issue of DNNs to some extent. 

Despite of encouraging progresses made in mitigating the spectral bias of DNNs, practical learning with DNNs remains a persistent challenge due to the bias, especially for learning from higher-dimensional data. This issue deserves further investigation. We propose to address this issue by understanding how a high-frequency function can be more accurately represented by neural networks. On one hand, it has been observed \cite{Cao2019, Rahaman2019, Ronen2019} that a function having only low frequency components can be well represented by a shallow neural network (SNN), a network having only a few layers. On the other hand, the classical Jacobi–Anger identity expresses a complex exponential of a trigonometric function as a linear combination of its harmonics that can contain significant high-frequency components. Even though the complex exponential function and the trigonometric function both are of low frequency, their composition could contain high frequency components. This motivates us to decompose a function containing high-frequencies as a {\it sum-composition} of low-frequency functions. That is, we decompose it into a sum of different frequency components, each of which is further broken down to a {\it composition of low-frequency functions}. In implementing this idea, we find that the multi-grade deep learning (MGDL) model recently introduced in  \cite{XuMultigrade2023first, XuSAL2023} matches seamlessly for constructing the sum-composition form for a function of high-frequency. It is the purpose of this study to introduce the general methodology in addressing the spectral bias issue of DNNs and implement it by employing MGDL as a technical tool. We demonstrate the efficacy of the proposed approach in four experiments with one-dimensional synthetic data, two-dimensional manifold data, two-dimensional colored images, and very high-dimensional modified National Institute of Standards and Technology (MNIST) data.
Our numerical results endorse that the proposed approach can effectively address the spectral bias issue, leading to substantial improvement in approximation accuracy in comparison with the traditional SGDL training approach.


\textbf{Contributions} of this paper include: (a) We propose a novel approach to address the spectral bias issue by decomposing a function containing high-frequencies as a sum of different frequency components, which are represented as compositions of low-frequency functions. (b) We investigate the efficacy of MGDL in decomposing a function of high-frequency into its ``sum-composition'' form of SNNs. (c) We successfully apply the proposed approach to synthetic data in 1 and 2 dimensions and real data in 2 and 784 dimensions,
showing that it can effectively address the spectral bias issue.  

\section{Proposed Approach and Multi-Grade Learning Model}\label{section: Frequency analysis}

We introduce a novel approach to tackle the spectral bias issue and review the MGDL model.



We begin with a quick review of the definition of DNNs. 
A DNN is a successive composition of an activation function composed with a linear transformation.  Let $\mathbb{R}$ denote the set of all real numbers, and $d, s$ be two positive integers. A DNN with depth $D$ consists an input layer, $D-1$ hidden layers, and an output layer. Let $\mathbb{N}_D:=\{1,2,\dots, D\}$. For $j\in\{0\}\cup\mathbb{N}_D$, let $d_j$ denote the number of neurons in the $j$-th hidden layer with $d_0:=d$ and $d_D:=s$. We use $\mathbf{W}_j \in \mathbb{R}^{d_j \times d_{j-1}}$ and $\mathbf{b}_j \in \mathbb{R}^{d_j}$ to represent the weight matrix and bias vector, respectively, for the $j$-th layer. By $\sigma: \mathbb{R} \to \mathbb{R}$ we denote an activation function.  When $\sigma$ is applied to a vector, it means that $\sigma$ is applied to the vector componentwise. For an input vector $\mathbf{x}:=[x_1, x_2, \ldots, x_d]^{\top} \in \mathbb{R}^d$,
the output of the first layer is defined by
$
\mathcal{H}_1(\mathbf{x}) := \sigma\left(  \mathbf{W}_{1}\mathbf{x}+\mathbf{b}_{1}\right).
$ 
For a DNN with depth $D\geq 3$, the output of the $(j+1)$-th hidden layer can be identified as a recursive function of the output of the $j$-th hidden layer, defined as
$
\mathcal{H}_{j+1}(\mathbf{x}) := \sigma\left(  \mathbf{W}_{j+1}\mathcal{H}_{j}\left(\mathbf{x}\right)+\mathbf{b}_{j+1}\right), 
$ 
for $j \in\mathbb{N}_{D-2}$. Finally, the output of the DNN with depth $D$ is an $s$-dimensional vector-valued function defined by
\begin{equation}\label{ND}
\mathcal{N}_{D}\left(\{\mathbf{W}_j,\mathbf{b}_j\}_{j=1}^D; \mathbf{x}\right) =\mathcal{N}_{D}(\mathbf{x}) :=    \mathbf{W}_{D}\mathcal{H}_{D-1}\left(\mathbf{x}\right)+\mathbf{b}_{D}.
\end{equation}
Suppose that data samples $\mathbb{D}:=\left\{\mathbf{x}_{\ell}, \mathbf{y}_{\ell}\right\}_{\ell = 1}^{N}$ are chosen. The loss 
on $\mathbb{D}$ is defined as
\begin{equation}\label{DNNs loss}
    \mathcal{L}\left(\left\{\mathbf{W}_j, \mathbf{b}_j\right\}_{j=1}^{D}; \mathbb{D}\right):= \frac{1}{2N}\sum\nolimits_{\ell=1}^{N}\norm{\mathbf{y}_{\ell} - \mathcal{N}_D\left(\left\{\mathbf{W}_j, \mathbf{b}_j\right\}_{j=1}^{D}; \cdot\right)(\mathbf{x}_{\ell})}_2^2.
\end{equation}
The traditional SGDL model is to minimize the loss function $L$ defined by \eqref{DNNs loss} with respect to $\Theta:=\{\mathbf{W}_j, \mathbf{b}_j\}_{j=1}^{D}$, which yields the optimal parameters $\Theta^*:=\{\mathbf{W}^*_j, \mathbf{b}^*_j\}_{j=1}^{D}$ and the corresponding DNN $\mathcal{N}_D\left(\Theta^*; \cdot\right)$.
When $D$ is relatively small, for example, $D<5$, we call $\mathcal{N}_{D}$ an SNN.
It is well-recognized that training an SNN is notably easier than training a DNN. 



\begin{SCfigure}
  \centering
   \begin{subfigure}{0.3\linewidth}
\includegraphics[width=\linewidth]{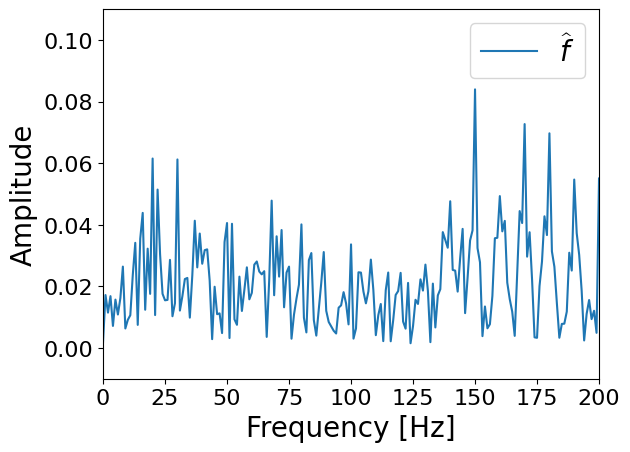}
   \end{subfigure}
   \begin{subfigure}{0.3\linewidth}
\includegraphics[width=\linewidth]{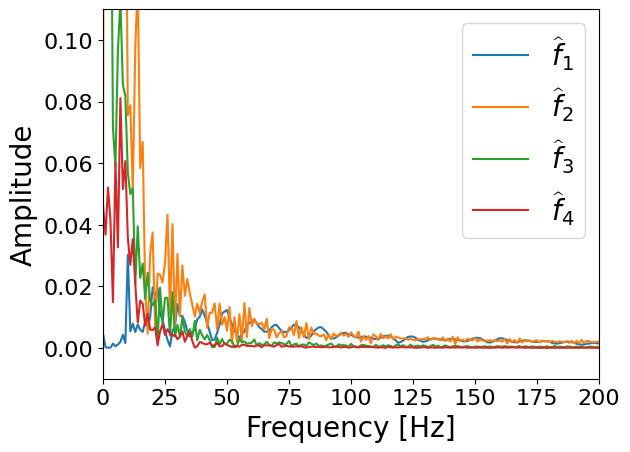}
    \end{subfigure}
	\caption{Spectrum comparison of $f:=f_1+f_2\circ f_1+f_3\circ f_2\circ f_1+f_4\circ f_3\circ f_2\circ f_1$ and $f_j$: Amplitude versus one-side frequency plots for function $f$ (Left) and $f_j$ for $j \in \mathbb{N}_{4}$ (Right). The function $f$ is of high-frequency and functions $f_j$ all are of low-frequency.}
	\label{fig: LearningviaComposition}
\end{SCfigure}

We motivate the proposed idea by a simple example.
We consider the function $f(\mathbf{x})$, $\mathbf{x}\in [0,1]$, whose Fourier transform is shown in Figure  \ref{fig: LearningviaComposition} (Left), where the Fourier transform is defined by $\hat{f}(\mathbf{t}) := \int_{-\infty}^{\infty} f(\mathbf{x})e^{-i2\pi\mathbf{t} \mathbf{x}} d\mathbf{x}$. To compute the Fourier transform of $f$ defined on $[0,1]$, we extend $f$ to the entire real line by assigning its value to be zero for $\mathbf{x}\notin [0,1]$. Observing from Figure  \ref{fig: LearningviaComposition} (Left), the function $f$ has significant high-frequency components, with frequencies varying from $0$ to $200$.
The function $f$ can be represented as
\begin{equation}\label{Decomposition}
    f(\mathbf{x}) = f_1(\mathbf{x})+(f_2\circ f_1)(\mathbf{x})+(f_3\circ f_2\circ f_1)(\mathbf{x})+(f_4\circ f_3\circ f_2\circ f_1)(\mathbf{x}), \ \ \mathbf{x}\in [0, 1],
\end{equation}
where $\circ$ denotes the composition of two functions. Note that the Fourier transforms $\hat{f}_j$, $j=1,2,3,4$, 
are displayed in Figure  \ref{fig: LearningviaComposition} (Right). Clearly, the functions $f_j$, $j=1,2,3,4$, are of low-frequency, with frequencies mainly concentrating on the interval $[0, 50]$. This example surely demonstrates that a function of high-frequency can be expressed as a sum of compositions of lower-frequency functions. This observation leads to the proposed approach of addressing the spectral bias of DNNs to be studied in this paper.
The legitimacy of the proposed idea may be reinforced by the Jacobi–Anger identity \cite{Arfken2011}, which expresses a complex exponential of a trigonometric function as a linear combination of its harmonics. Even though both the complex exponential function and the trigonometric function are of low-frequency, their composition contains many high-frequency components. 
We now review the Jacobi–Anger identity,
the identity named after the 19th-century mathematicians Carl Jacobi and Carl Theodor Anger. It has the form
\vspace{-2mm}
\begin{equation}\label{Jacobi–Anger identity}
    e^{ia\sin( b\mathbf{x})} = \sum\nolimits_{n = -\infty}^{\infty} J_n(a) e^{i nb \mathbf{x}},
\end{equation}
where $i$ denotes the imaginary unit and $J_n(a)$ denotes the $n$-th Bessel function of the first kind, see details in \cite{Arfken2011}.
Taking the real part of the both sides of the  Jacobi–Anger identity \eqref{Jacobi–Anger identity}, we obtain that
\begin{equation}\label{composition of low frequency component}
\cos(a \sin(b\mathbf{x}) ) = \sum\nolimits_{n=-\infty}^{\infty}J_{n}(a)\cos(n b\mathbf{x}).
\end{equation}
The left-hand side of \eqref{composition of low frequency component} is a composition of two low-frequency functions $\cos(a \mathbf{x})$ and $\sin(b\mathbf{x})$, having frequencies $a/(2\pi)$ and $b/(2\pi)$, respectively, while the right-hand side 
is a linear combination of $\cos(n b\mathbf{x})$ with $n$ taking all integers. The high-frequency of the composition can be estimated by a rule of thumb.
Specifically, the left-hand side of \eqref{composition of low frequency component} is a frequency-modulated sinusoidal signal \cite{Schulze2003, Shanmugam1979}, with its frequencies spreading on an interval centered at zero. 
It follows from the well-known Carson bandwidth rule \cite{Carson1922, Pieper2011, Schulze2003}, regarded as a rule of thumb, that more than $98\%$ frequencies are located within the interval $[-(ab+b)/(2\pi), (ab+b)/(2\pi)]$. Therefore, the highest frequency of $\cos(a \sin(b\mathbf{x}))$ can be well-estimated by $(ab+b)/(2\pi)$, which is greater than the product of the frequencies of $\cos(a\mathbf{x})$ and $\sin(b\mathbf{x})$. These suggest that a composition of two low-frequency functions may lead to a high-frequency function.

The example presented earlier, together with the Jacobi–Anger identity, inspires us to decompose a given function into a sum of different frequency components, each of which is a composition of lower-frequency functions, a decomposition similar to equation \eqref{Decomposition} for the function $f$ represented in Figure \ref{fig: LearningviaComposition} (Left). In other words, for a function $g$ of high-frequency, we decompose it in a ``sum-composition'' form as
\begin{equation}\label{Decomposition-General}
    g=\sum\nolimits_{k=1}^K \bigodot\nolimits_{j=1}^k g_j,
\end{equation}
where
$\bigodot_{j=1}^k g_j:=g_k\circ\cdots\circ g_2\circ g_1
$,  and $g_j$, $j\in\mathbb{N}_k$, are all of low-frequency.
The function $f$ represented in \eqref{Decomposition} is a special example of \eqref{Decomposition-General}. In the context of approximation by neural networks, we prefer expressing $g_j$ by SNNs, as a function having only low-frequency components can be well-represented by an SNN. The MGDL model originated in \cite{XuMultigrade2023first, XuSAL2023} furnishes exactly the decomposition \eqref{Decomposition-General}, with each $g_j$ being an SNN. We propose to employ MGDL to learn the decomposition \eqref{Decomposition-General}, where the low-frequency function $g_j$ is represented by an SNN.

It is worth explaining the motivation behind the MGDL model. MGDL was inspired by the human education system which is arranged in grades. In such a system, students learn a complex subject in grades, by decomposing it into sequential, simpler topics. Foundational knowledge learned in previous grades remains relatively stable and serves as a basis for learning in a present and future grades. This learning process can be modeled mathematically by representing a function that contains higher-frequency components by a ``sum-composition'' form of low-frequency functions. MGDL draws upon this concept by decomposing the learning process into multiple grades, where each grade captures different levels of complexity. 

We now review the MGDL model that learns given data $\mathbb{D}:=\left\{\mathbf{x}_{\ell}, \mathbf{y}_{\ell}\right\}_{\ell=1}^{N}$. Following \cite{XuMultigrade2023first}, we split a DNN with depth $D$ into $L$ grades, with $L<D$, each of which learns an SNN  $\mathcal{N}_{D_l}$, defined as \eqref{ND}, with depth $D_l$, from the residue $\{\mathbf{e}_{\ell}^l\}_{\ell=1}^N$ of the previous grade, where $1<D_l<D$ and $\sum_{l=1}^{L}D_{l} = D+L-1$. Let $\Theta_l:=\left\{\mathbf{W}^l_j, \mathbf{b}^l_j\right\}_{j=1}^{D_l}$ denote the parameters to be learned in grade $l$. We define recursively  $g_1(\Theta_1;\mathbf{x}):= \mathcal{N}_{D_1}\left(\Theta_1; \mathbf{x}\right)$, $g_{l+1}(\Theta_{l+1}; \mathbf{x}):= \mathcal{N}_{D_{l+1}}\left(\Theta_{l+1};\cdot\right) \circ \mathcal{H}_{D_{l}-1}(\Theta_{l}^*;\cdot) \circ \ldots \circ \mathcal{H}_{D_{1}-1}(\Theta_1^*; \cdot)(\mathbf{x})$, for $l\in\mathbb{N}_{L-1}$, and the loss function of grade $l$ by
\vspace{-2mm}
\begin{equation}\label{Loss:Grade_l}
    \mathcal{L}_l\left(\Theta_l; \mathbb{D}\right):= \frac{1}{2N}\sum\nolimits_{\ell=1}^{N}\norm{\mathbf{e}_{\ell}^l - g_l\left(\Theta_l; \mathbf{x}_{\ell}\right)}_2^2,
\end{equation} 
\vspace{-3mm}

\noindent
where $\Theta_l^*:=\left\{\mathbf{W}^{l*}_j, \mathbf{b}^{l*}_j\right\}_{j=1}^{D_l}$ are the  optimal parameters learned by minimizing the loss function $\mathcal{L}_l$ with respect to $\Theta_l$.
The residues are defined by
 $\mathbf{e}_{\ell}^{1}:= \mathbf{y}_{\ell}$ and  $\mathbf{e}^{l+1}_{\ell}:= \mathbf{e}^{l}_{\ell} - g_{l}(\Theta_{l}^*;\mathbf{x}_{\ell})$, for $l\in \mathbb{N}_{L-1}$, $\ell \in \mathbb{N}_{N}$.
When minimizing the loss function $\mathcal{L}_l(\Theta_l; \mathbb{D})$ of grade $l$, parameters $\Theta^*_j$, $j\in \mathbb{N}_{l-1}$, learned from the previous $l-1$ grades are all {\it fixed} and $\mathcal{H}_{D_{l-1}-1}(\Theta_{l-1}^*;\cdot) \circ \ldots \circ \mathcal{H}_{D_{1}-1}(\Theta_1^*; \cdot)$ serves as a feature or ``basis".
After $L$ grades are learned, the function $\bar{g}_{L}$ learned from MGDL is the summation of the function learned in each grade, that is,
\begin{equation}\label{composition}
\bar{g}_L\left(\left\{\Theta^*_{l}\right\}_{l=1}^L; \mathbf{x}\right):= \sum\nolimits_{l=1}^{L}g_l(\Theta_{l}^{*}; \mathbf{x}),   
\end{equation}
where
$g_{l}(\Theta_{l}^*; \mathbf{x}):= \mathcal{N}_{D_{l}}\left(\Theta^*_{l};\cdot\right) \circ \mathcal{H}_{D_{l-1}-1}(\Theta_{l-1}^*;\cdot) \circ \ldots \circ \mathcal{H}_{D_{1}-1}(\Theta_1^*; \cdot)(\mathbf{x})$, and 
$\mathcal{H}_{D_{k-1}-1}$ for $1\leq k \leq L$ and $\mathcal{N}_{D_{L}}$ are SNNs learned in different grades. Thus, MGDL enables us to construct the desired ``sum-composition" form \eqref{Decomposition-General}. When $L=1$, MGDL reduces to the traditional SGDL model.

In MGDL, we use the mean squared error (MSE) loss function. It was established in \cite{XuMultigrade2023first} that when the loss function is defined by MSE, MGDL either learns the zero function or results in a strictly decreasing residual error sequence (see, Theorem \ref{Theorem_1} in Appendix \ref{AnalysisofMGDL}).
Since the regression problems conducted in this paper naturally align with MSE losses, it is a suitable choice. In practice, MGDL can also be applied with other loss functions, such as cross-entropy loss, when solving classification problems.
In MGDL, the computation cost remains relatively consistent across all grades. For
$
\mathbf{x}_{\ell}^{l} :=\mathcal{H}_{D_{l-1}-1}(\Theta_{l-1} 
        ^*;\cdot) \circ \mathcal{H}_{D_{l-2}-1}(\Theta_{l-2} 
        ^*;\cdot)\circ \ldots \circ \mathcal{H}_{D_{1}-1}(\Theta_{1}^*;\cdot)(\mathbf{x}_{\ell}),
$
we recursively let $
 \mathbf{x}_{\ell}^1:= \mathbf{x}_{\ell}, \ \ \mathbf{x}_{\ell}^{k} :=\mathcal{H}_{D_{k-1}-1}(\Theta_{k-1} 
    ^*;\cdot) \circ \mathbf{x}_{\ell}^{k-1}, \ \ k=2, 3, \dots, n.$
When training grade $l$, we use the output of grade $l-1$, denoted as $\mathbf{x}_{\ell}^{l}$ along the residual $\mathbf{e}_{\ell}^{l}$, which are already obtained.
The training dataset in grade $l$ consists of $\left\{(\mathbf{x}_{\ell}^{l}, \mathbf{e}_{\ell}^{l})\right\}_{\ell=1}^{N}$. This dataset is used to train a new shallow network, which is independent of the previous $l-1$ grades.  Moreover, $\mathbf{x}_{\ell}^{l}$ can be computed recursively, ensuring that the computation cost for each grade remains relatively consistent.

MGDL avoids training a DNN from end to end.  Instead, it trains several SNNs sequentially, with the current grade making use the SNNs learned from the previous grades as a feature and composing it with a new SNN to learn the residue of the previous grade. This allows MGDL to decompose a function that contains higher-frequency in a form of \eqref{Decomposition-General}, with $g_j$ being a SNN learned from grade $j$. In this way, higher-frequency components in the data can be effectively learned in a grade-by-grade manner. Note that the training time of MGDL increases linearly with the number of grades. This makes MGDL an effective and scalable solution for tackling complex tasks. MGDL is an adaptive approach by nature. When the outcome of the present grade is not satisfactory, we can always add a new grade without changing the previous grades.

It is worth noting that while ResNet \cite{He2016} also has a sum-composition form, MGDL differs from it significantly. The ``Composition'' for ResNet refers to composition of layers, while that for MGDL emphasizes the composition of the SNNs sequentially learned in the previous grades. Moreover, ResNet learns all parameters of the entire sum of DNNs at once, training it from end to end, whereas MGDL learns the sum incrementally, grade by grade, in each grade training an SNN composed with the feature (the composition of the SNNs learned in the previous grades).
MGDL also differs from the relay backpropagation approach proposed in \cite{ShenLinHuang2016}, where a DNN is divided into multiple segments, each with its own loss function. The gradients from these losses are then propagated to lower layers of their respective segments and all segments are optimized all together by minimizing the sum of the losses. While MGDL trains SNNs in a multi-grade manner, each of which learns from the residue of the previous grade, freezing the previously learned SNNs (serving as features or ``bases'').

MGDL is a principle applicable to various models, including standard DNNs, convolutional neural networks, and ResNet. In this paper, we demonstrate its feasibility by applying it to standard DNNs.

\vspace{-3mm}
\section{Numerical Experiments}\label{numerical experiments}
\vspace{-3mm}

In this section, we study MGDL empirically in addressing the spectral bias issue of SGDL. We consider four examples:
Subsections 3.1, 3.2, and 3.4 investigate regression on synthetic, manifold, and MNIST data, respectively, for which the spectral bias phenomena of SGDL are identified in \cite{Rahaman2019}. Section 3.3 deals with regression on colored images, which were studied in \cite{Tancik2020} by using the Fourier features to mitigate the spectral bias. Our goal is to compare the learning performance of MGDL with that of SGDL on these datasets and understand to what extent MGDL can overcome the spectral bias exhibited in SGDL.

The loss functions defined in \eqref{DNNs loss} for SGDL and \eqref{Loss:Grade_l} for MGDL are used to compute the training and validation loss when $\mathbb{D}$ is chosen to be the training and validation data, respectively.
We use the relative squared error (RSE) to measure the accuracy of predictions obtained from both SGDL and MGDL. Assume that $\mathcal{N}$ is a trained neural network. For a prediction value $\hat{\mathbf{y}}_{\ell}:=\mathcal{N}(\mathbf{x}_{\ell})$ at $\mathbf{x}_{\ell}$, we define
$
\text{RSE}:= 
{\sum_{\ell = 1}^N\norm{\hat{\mathbf{y}}_{\ell} - \mathbf{y}_{\ell}}_2^2}/{\sum_{\ell = 1}^N \norm{\mathbf{y}_{\ell}}_2^2}
$. When $\mathbb{D}$ represents the training, validation, and testing data, RSE 
is specialized as TrRSE, VaRSE, and TeRSE, respectively.

Details of the numerical experiments conducted in this section, including computational resources, the network structure of SGDL and MGDL for each example, the choice of activation function, the optimizer, parameters used in the optimization process, and supporting figures are provided in Appendix \ref{Experimental details}. 


{\bf 3.1  Regression on the synthetic data.} 
In this experiment, we compare the efficacy of  SGDL and MGDL in learning functions of four different types of high-frequencies. 


The experiment setup is as follows. Given frequencies $\kappa:=\left( \kappa_1, \kappa_2, \ldots, \kappa_{M}\right)$ with corresponding amplitudes $\alpha:= \left(\alpha_1, \alpha_2, \ldots, \alpha_{M}\right)$, and phases $\varphi:=\left(\varphi_1, \varphi_2, \ldots, \varphi_M\right)$, we consider approximating the function $\lambda:[0,1] \rightarrow \mathbb{R}$ defined by
\vspace{-2mm}
\begin{equation}\label{regression function}
    \lambda(\mathbf{x}) := \sum\nolimits_{j=1}^{M} \alpha_j \sin\left(2 \pi \kappa_j \mathbf{x} + \varphi_j\right), \quad \mathbf{x} \in [0,1] 
\end{equation}
by neural networks learned with SGDL and MGDL. 
We consider four settings, in all of which we choose $M:=20$, $\kappa_j := 10j$ and $\varphi_j \sim \mathcal{U}(0,2 \pi)$ for $j \in \mathbb{N}_{20}$, where $\mathcal{U}$ denotes the uniform distribution and the random seed is set to be $0$. In settings $1$, $2$, $3$, and $4$, we choose respectively the amplitudes $\alpha_j:=1$,  $\alpha_j := 1.05-0.05j$, $\alpha_j(\mathbf{x}):= e^{-\mathbf{x}}\cos(j\mathbf{x})$, and $\alpha_j := 0.05j$,  for $j \in \mathbb{N}_{20}$. Note that in setting $3$, we explore the case where the amplitude varies as a function of $\mathbf{x}$ for each component. 
The amplitudes versus one-side frequencies for $\lambda$ across the four settings are depicted in Figure \ref{fig: frequency constant increase vary original function} in Appendix \ref{Experimental details 3.1}. 
For all the four settings, the training data consist of pairs $\left\{\mathbf{x}_{\ell},  \lambda(\mathbf{x}_{\ell})\right\}_{\ell\in \mathbb{N}_{6000}}$, where $\mathbf{x}_{\ell}$'s are equally spaced between $0$ and $1$. The validation and testing data consist of pairs $\left\{\mathbf{x}_{\ell},  \lambda(\mathbf{x}_{\ell})\right\}_{\ell\in \mathbb{N}_{2000}}$, where $\mathbf{x}_{\ell}$'s are generated from a random uniform distribution on $[0, 1]$, with the random seed set to be 0 and 1, respectively.

\begin{figure}
  \centering
   \begin{subfigure}{0.24\linewidth}  \includegraphics[width=\linewidth]{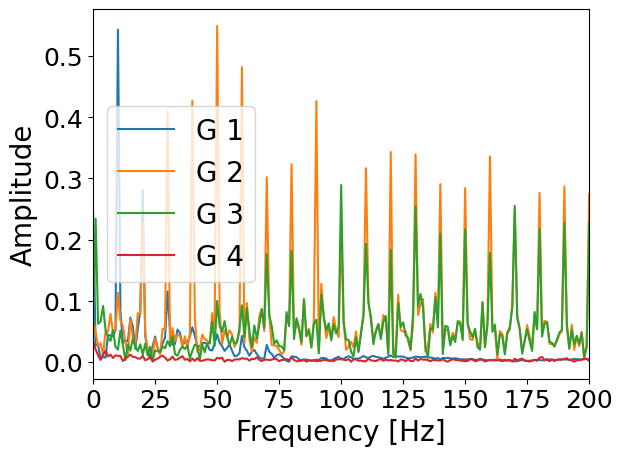}
   \end{subfigure}
   \begin{subfigure}{0.24\linewidth}  \includegraphics[width=\linewidth]{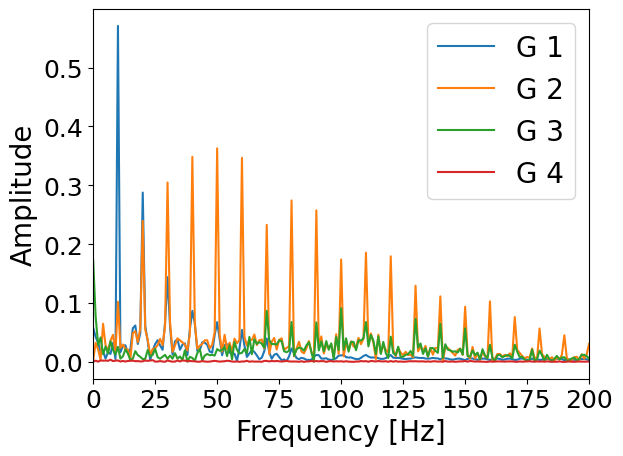}
   \end{subfigure}
   \begin{subfigure}{0.24\linewidth}
        \includegraphics[width=\linewidth]{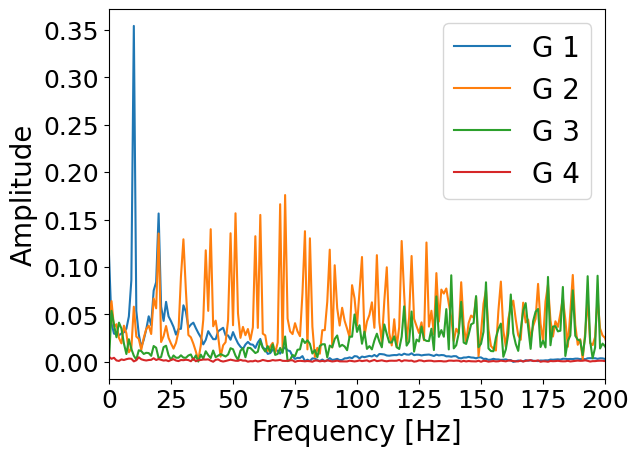}
    \end{subfigure}
   \begin{subfigure}{0.24\linewidth}
        \includegraphics[width=\linewidth]{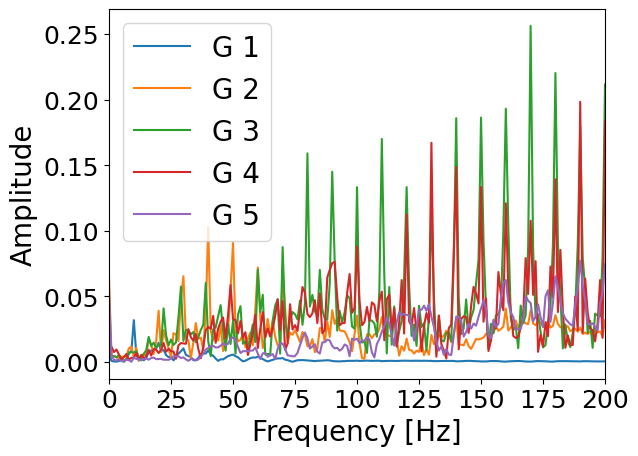}
    \end{subfigure}
	\caption{Amplitude versus one-side frequency plot for the learned functions learned across four
grades of MGDL for settings 1-4.
}
	\label{fig: learned frequency constant increase vary}
\end{figure}

\begin{figure}[ht]
    \centering
    \begin{minipage}{0.49\linewidth}
        \centering
        \includegraphics[width=0.49\linewidth]{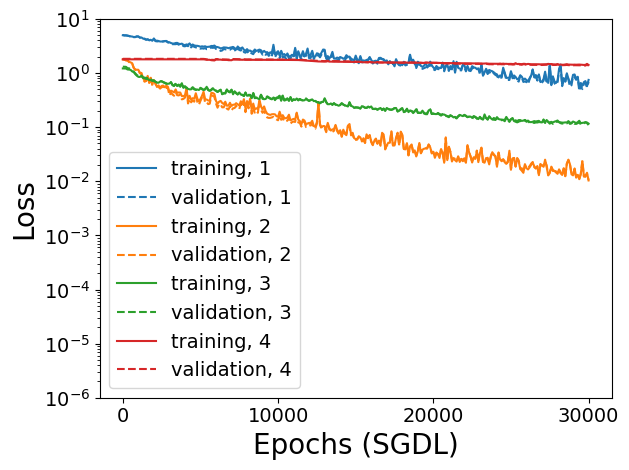}
        \includegraphics[width=0.49\linewidth]{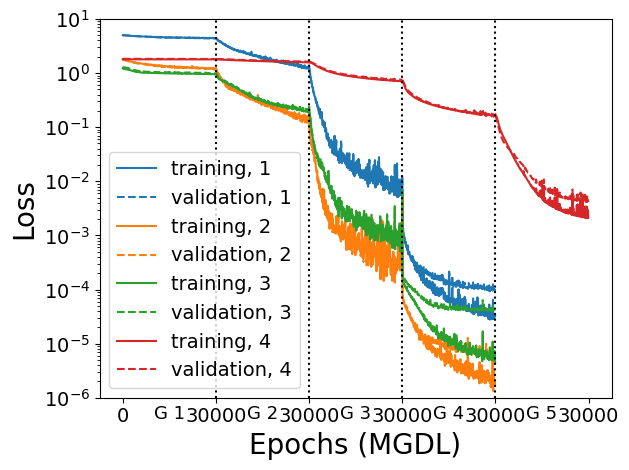}
        \caption{Comparison of SGDL (left) and MGDL (right): training and validation loss across settings 1-4.}
        \label{fig: loss compare SDNN and MDNN}
    \end{minipage}
    \hfill
    \begin{minipage}{0.49\linewidth}
        \centering
        \includegraphics[width=0.49\linewidth]{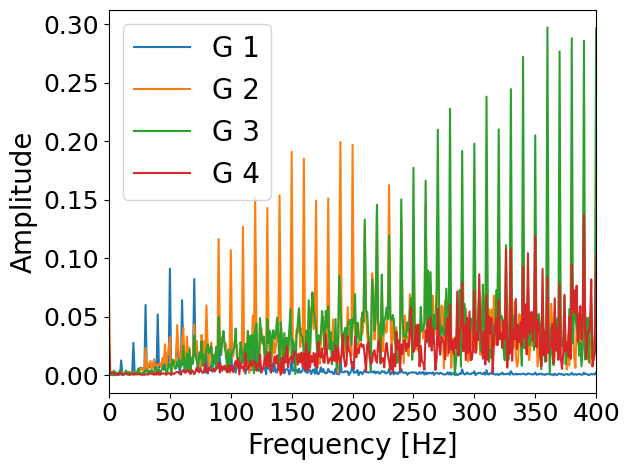}
        \includegraphics[width=0.49\linewidth]{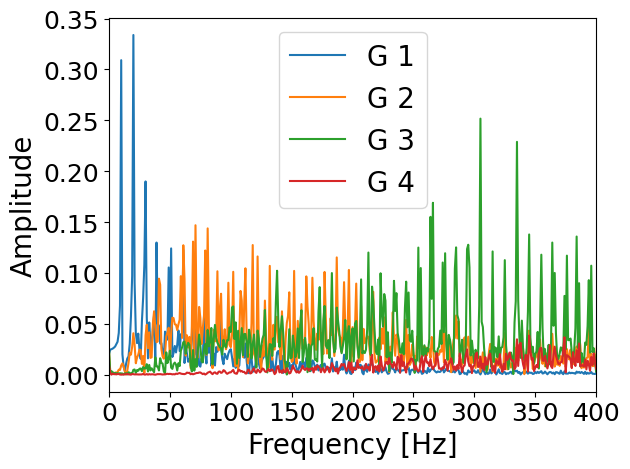}
        \caption{Amplitude vs. one-side frequency plot for the learned function across four grades of MGDL: settings 1 (left) and 2 (right) with $q=0$.}
        \label{fig: learned frequency for each grade MDNN setting 1 and 2}
    \end{minipage}
\end{figure}

Numerical results for this example are reported in Figures  \ref{fig: learned frequency constant increase vary} and \ref{fig: loss compare SDNN and MDNN}, as well as \ref{fig: frequency constant increase vary} and \ref{fig: spetral compare SDNN and MDNN} (in Appendix \ref{Experimental details 3.1}), and Table \ref{tab: Spectral constant increase vary}. 
Figure \ref{fig: learned frequency constant increase vary} displays the amplitude versus the one-side frequency of the functions learned across four grades of MGDL in  settings 1-3, and five grades in setting 4.  In all the four settings, MGDL exhibits a pattern of learning low-frequency components in grade 1, middle-frequency components in grade 2, and high-frequency components in grades 3, 4 (and 5 for setting 4). We let $\mathcal{N}^*_j:=\mathcal{N}_{D_j}(\Theta^*_j;\cdot)$ and $\mathcal{H}^*_j:=\mathcal{H}_{D_j-1}(\Theta^*_j;\cdot)$.
The SNN $\mathcal{N}^*_1$ learned in grade 1 represents a low-frequency component of $\lambda$, the SNNs  $\mathcal{N}^*_2$ and $\mathcal{N}^*_3$ learned in grades 2 and 3, composed with $\mathcal{H}^*_1$ and $\mathcal{H}^*_2\circ\mathcal{H}^*_1$, respectively, represents higher-frequency components of $\lambda$.  Likewise, the SNN $\mathcal{N}^*_4$, learned in grade 4, composed with $\mathcal{H}^*_3\circ\mathcal{H}^*_2\circ\mathcal{H}^*_1$ represents the highest-frequency component of $\lambda$. 
This fact is particularly evident in setting 4 (see, the fourth subfigure in Figure \ref{fig: learned frequency constant increase vary}), where the amplitude within the data increases with the frequency. For settings 1, 2, and 3, grade 4 does not learn much. This is because for the functions of these three settings, the amplitudes of higher-frequencies are {\it proportionally} small. However, grade 4 is still important for these settings. As shown in Figure \ref{fig: loss compare SDNN and MDNN} (right), grade 4 reduces the loss from $10^{-2}$ to $10^{-5}$ for setting 1 and from $10^{-4}$ to $10^{-6}$ for settings 2 and 3. This indicates that if we want a high precision, we need to include grade 4. For setting 4, we need grade 5 to learn its highest frequency component.
These findings suggest that MGDL is particularly well-suited for learning the high-frequency components of the function.

\begin{small}
\begin{table}[]
    \centering
     \caption{Comparison of  SGDL and MGDL: Accuracy}
    \resizebox{\textwidth}{!}{
    \begin{tabular}{c||c|cccrccc}
    \hline
     setting&  model & $t_{max}$ & $t_{min}$ & batch size& time(s) & TrRSE   & VaRSE & TeRSE\\
\hline
\multirow{2}{*}{ 1 } 
 & SGDL & $10^{-4}$& $10^{-4}$ & $256$& $32,401$& $1.5 \times 10^{-1}$ & $1.3 \times 10^{-1}$ & $1.2\times 10^{-1}$
\\
& 
 MGDL & $10^{-4}$ & $10^{-4}$ & $256$&  $27,817$ & $\mathbf{5.9 \times 10^{-6}}$ & $\mathbf{1.9 \times 10^{-5}}$ & $\mathbf{1.7\times 10^{-5}}$
\\
\hline
\multirow{2}{*}{ 2 } 
 & SGDL & $5\times 10^{-4}$& $10^{-4}$ & $256$& $32,053$ & $5.8 \times 10^{-3}$ & $5.9 \times 10^{-3}$ & $5.7\times 10^{-3}$
\\
& 
 MGDL & $10^{-4}$ & $10^{-4}$ & $256$ &  $27,628$ & $\mathbf{1.5 \times 10^{-6}}$ & $\mathbf{4.0 \times 10^{-6}}$ & $\mathbf{6.5\times 10^{-6}}$
\\
\hline
 \multirow{ 2}{*}{ 3 }
& SGDL & $10^{-3}$ & $10^{-4}$ & $256$ & 
$31,808$ & $9.6\times 10^{-2}$ & $8.3\times 10^{-2}$ & $1.1 \times 10^{-1}$
\\
 & MGDL & $10^{-4}$ & $10^{-4}$ & $256$ &  $24,876$ & $\mathbf{5.2 \times 10^{-6}}$ & $\mathbf{3.2 \times 10^{-5}}$ & $\mathbf{2.1\times 10^{-5}}$
\\
\hline
\multirow{ 2}{*}{ 4 }
& SGDL & $5\times 10^{-5}$ & $10^{-5}$ & $256$ & $41,063$ & $7.9\times 10^{-1}$ & $7.5 \times 10^{-1}$ & $7.7\times 10^{-1}$
\\
& MGDL & $10^{-4}$ & $10^{-4}$ & Full & $9,875$ & $\mathbf{1.1 \times 10^{-3}}$ & $\mathbf{2.2 \times 10^{-3}}$ & $\mathbf{1.3\times 10^{-3}}$
\\
\hline
    \end{tabular}
    }
    \label{tab: Spectral constant increase vary}
\end{table}
\end{small}

Figure \ref{fig: loss compare SDNN and MDNN} compares the progress of the training and validation losses against the number of training epochs for SGDL and MGDL across the four settings. We observe that when learning a task involving high-frequency components by SGDL, the training and validation losses decrease slowly due to the spectral bias of DNN. While the same task is learned by MGDL, the learning process progresses through distinct grades. In grade 1, MGDL primarily learns low-frequency, resulting in a slow decrease in loss. In grade 2, the training loss and validation loss both decrease more rapidly due to the use of the composition of SNN $\mathcal{N}^*_2$ with the feature $\mathcal{H}^*_1$, facilitating in learning high-frequency features. This accelerated learning aspect of MGDL is further evidenced in grades 3 and 4 (as well as grade 5 for setting 4). Table \ref{tab: Spectral constant increase vary} compares the accuracy achieved by SGDL and MGDL. 
Within a comparable or even less training time, MGDL increases accuracy, measured by TeRSE from $10^{-1}$ to $10^{-5}$, $10^{-3}$ to $10^{-6}$,  $10^{-1}$ to $10^{-5}$, and $10^{-1}$ to $10^{-3}$ in settings 1, 2, 3 and 4, respectively. Across the four settings, TeRSE values are reduced by a factor of $592\sim 7,058$. 
These comparisons highlight  MGDL's advantage in effectively learning high-frequency oscillatory functions.

Figure \ref{fig: frequency constant increase vary} in Appendix \ref{Experimental details 3.1} depicts the functions, in the Fourier domain, learned by SGDL (row 1) and MGDL (row 2) across the four settings, 
demonstrating that MGDL has a substantial reduction in the `spectral bias' exhibited in SGDL.  
This is because high-frequency components are learned in a higher grade, where they are represented as the composition of a low-frequency component with the low-frequency components learned in the previous grades,
and each grade focuses solely on learning a low-frequency component by an SNN. 
We also include Figure \ref{fig: spetral compare SDNN and MDNN} in Appendix \ref{Experimental details 3.1} to compare the spectrum evolution between SGDL (1st row) and MGDL (2nd row) across settings 1-4.
Notably, although in iterations SGDL and MGDL both learn low-frequency components first and then followed by middle and high-frequency components, MGDL learns in grade by grade, exhibiting significant outperformance.


{\bf 3.2 Regression on the manifold data.}
The second experiment compares regression by SGDL and MGDL on  
two-dimensional manifold data, studied in \cite{Rahaman2019} but with twice higher frequencies. 

The goal of this experiment is to explore scenarios where data lies on a lower-dimensional manifold embedded within a higher-dimensional space. Such data is commonly referred to as manifold data \cite{Bengio2012RepresentationLA}. 
Let $\gamma$ be an injective mapping from $[0, 1]^m \to \bR^d$ with $m \leq d$ and $\mathcal{M}:=\gamma([0, 1]^m)$ denote the manifold data. 
A  target function $\tau: \mathcal{M} \to \bR$ defined on the manifold can be identified with function $\lambda := \tau \circ \gamma$ defined on $[0, 1]^m$. Regressing the target function $\tau$ is therefore equivalent to finding $f: \bR^d \to \bR$ such that $f \circ \gamma$ matches $\lambda$.
Following \cite{Rahaman2019}, we set $m:=1$, $d:=2$, and choose 
the mapping $\gamma$ as 
\begin{minipage}{\linewidth}
\begin{equation}\label{manifold: gamma}
\gamma_q(\mathbf{x}) := \left[1+\sin(2\pi q \mathbf{x})/2\right]\left(\cos(2\pi \mathbf{x}), \sin(2\pi \mathbf{x})\right), \ \ \mathbf{x}\in [0, 1],   \end{equation}
\end{minipage}
for a nonnegative integer $q$. Clearly, $\gamma_q: [0, 1] \to \bR^2$, and $\mathcal{M} :=\gamma_q([0, 1])$ defines the manifold corresponding to a flower-shaped curve with $q$ petals when $q>0$, and a unit circle when $q=0$. Suppose that $\lambda: [0, 1] \to \bR$ is the function defined by \eqref{regression function}. Our task is to learn a DNN $f: \bR^2 \to \bR$ such that $f\circ \gamma_q$ matches $\lambda$. 
We consider two settings for $\lambda$. In settings 1 and 2, we choose $\alpha_j := 0.025 j$ and $\alpha_j(\mathbf{x}):= e^{-\mathbf{x}}\cos(j\mathbf{x})$ for $j \in \mathbb{N}_{40}$, respectively. For both the settings, we choose $\kappa_j := 10 j$ and $\varphi_j \sim \mathcal{U}(0, 2\pi)$ for $j \in \mathbb{N}_{40}$ with the random seed set to be $0$, and consider the cases where $q:=4$ and $q:=0$.  Note that the smaller $q$ is, the more difficult the learning task is.
The training data consists of pairs $\left\{\gamma_q(\mathbf{x}_{\ell}),  \lambda(\mathbf{x}_{\ell})\right\}_{\ell\in \mathbb{N}_{12000}}$, where $\mathbf{x}_{\ell}$'s are equally spaced between $0$ and $1$. The validation and testing data consist of pairs $\left\{\gamma_q(\mathbf{x}_{\ell}),  \lambda(\mathbf{x}_{\ell})\right\}_{\ell \in \mathbb{N}_{4000}}$, where $\mathbf{x}_{\ell}$'s are generated from a random uniform distribution on $[0, 1]$, with random seed set to be 0 and 1, respectively.

Numerical results for this example are reported in Figures \ref{fig: learned frequency for each grade MDNN setting 1 and 2}-\ref{fig: compare MDNN and SDNN setting loss}, and \ref{fig: compare MDNN and SDNN setting 1} (in Appendix \ref{Experimental details 3.2}), and Table \ref{tab: Manifold}. Figure \ref{fig: learned frequency for each grade MDNN setting 1 and 2} illustrates the frequency of functions learned across four grades of MGDL for settings 1 and 2, where $q:=0$. In both of the settings, MGDL exhibits a pattern of learning low-frequency components in grade $1$, middle-frequency components in grade $2$, and high-frequency components in grades $3$ and $4$. Therefore, the high-frequency components within the function mainly learned in higher grades, in which the learned function is a composition of the SNNs learned from several grades. That is, MGDL decomposes a high-frequency component as the composition of several lower-frequency components, facilitating effectively learning high frequency features within the data.

\begin{small}

\begin{table}[]
    \centering
    \caption{Accuracy comparison: SGDL versus MGDL.}
    \resizebox{\textwidth}{!}{
    \begin{tabular}{c||c|c|cccrccc}
    \hline
    setting & q & method & $t_{max}$ & $t_{min}$ & batch size& time (s) & TrRSE  &VaRSE  &TeRSE\\
\hline
\multirow{4}{*}{ 1 } 
& \multirow{ 2}{*}{ 4 } & SGDL & $10^{-4}$ & $10^{-6}$ & $1024$ & $28,832$ & $2.7\times 10^{-1}$ & $2.4 \times 10^{-1}$ & $2.8\times 10^{-1}$
\\
& & MGDL & $10^{-3}$ & $10^{-4}$ & Full &  $15,519$ & $\mathbf{4.9 \times 10^{-5}}$ & $\mathbf{1.8 \times 10^{-4}}$ & $\mathbf{1.1\times 10^{-4}}$
\\
  \cdashline{2-10}[1pt/1pt]
& \multirow{ 2}{*}{ 0 } & SGDL & $10^{-4}$ & $10^{-5}$ & 
$1024$ & $29,051$ & $5.5\times 10^{-1}$ & $5.4\times 10^{-1}$ & $5.4 \times 10^{-1}$
\\
& & MGDL & $10^{-3}$ & $10^{-4}$ & Full &  $15,969$ & $\mathbf{1.9 \times 10^{-4}}$ & $\mathbf{2.7 \times 10^{-4}}$ & $\mathbf{2.1\times 10^{-4}}$
\\
\hline   
\multirow{4}{*}{ 2 }

& \multirow{ 2}{*}{ 4 } & SGDL & $10^{-3}$ & $10^{-6}$ & $512$ & $44,083$ & $2.0\times 10^{-4}$ & $1.8 \times 10^{-3}$ & $2.3\times 10^{-4}$
\\
& & MGDL & $10^{-3}$ & $10^{-4}$ & Full &  $11,067$ & $\mathbf{8.5 \times 10^{-6}}$ & $\mathbf{1.4 \times 10^{-3}}$ & $\mathbf{4.2\times 10^{-5}}$

\\   
  \cdashline{2-10}[1pt/1pt]
&  \multirow{ 2}{*}{ 0 } & SGDL & $10^{-4}$ & $10^{-4}$ & $512$ & $41,941$ & $1.1\times 10^{-2}$ & $1.8\times 10^{-2}$ & $1.0 \times 10^{-2}$
\\
& & MGDL & $10^{-3}$ & $10^{-4}$ & Full &  $11,027$ & $\mathbf{3.7 \times 10^{-5}}$ & $\mathbf{2.2 \times 10^{-3}}$ & $\mathbf{8.7\times 10^{-5}}$
\\
\hline
    \end{tabular}
}
    \label{tab: Manifold}
\end{table}
\end{small}

Table \ref{tab: Manifold} compares the approximation accuracy achieved by SGDL and MGDL for settings 1 and 2. For SGDL, reducing the value of $q$ makes the learning task for both settings more challenging, due to the spectral bias of DNNs. 
 When $q:=4, 0$ in setting 1 and $q:=0$ in setting 2, learning becomes especially challenging for SGDL. In such cases, MGDL significantly outperforms SGDL by achieving higher accuracy in approximately half to one-third of the training time for both settings.  Figure \ref{fig: compare MDNN and SDNN setting loss} displays the training and validation loss for SGDL and MGDL.
 Figure \ref{fig: compare MDNN and SDNN setting 1} illustrates the spectrum evolution throughout the learning process for settings 1 and 2.  We observe that MGDL outperforms SGDL in accuracy  during the learning process. 
 Table \ref{tab: Manifold} and Figure \ref{fig: compare MDNN and SDNN setting 1} consistently demonstrate that MGDL exhibits a substantial advancement in addressing the spectral bias.

\begin{figure}
  \centering
    \begin{subfigure}{0.24\linewidth}
        \includegraphics[width=\linewidth]{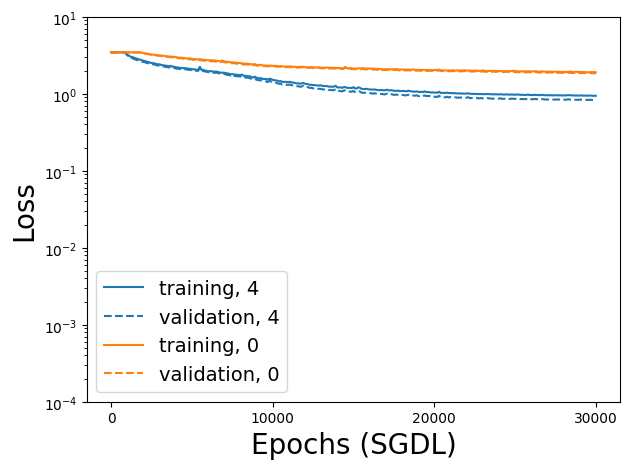}
    \end{subfigure} 
   \begin{subfigure}{0.24\linewidth}
        \includegraphics[width=\linewidth]{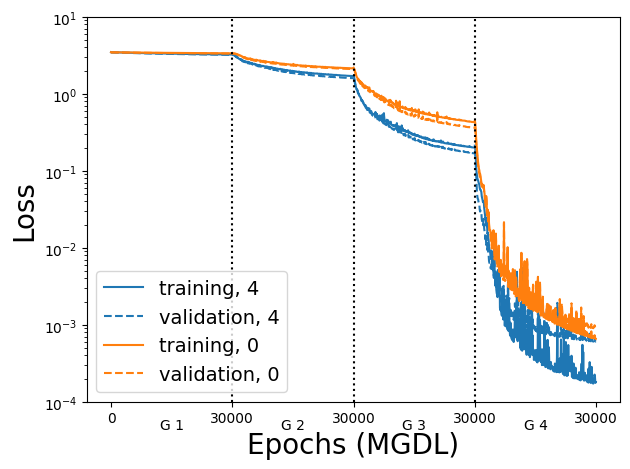}
   \end{subfigure}
    \begin{subfigure}{0.24\linewidth}
        \includegraphics[width=\linewidth]{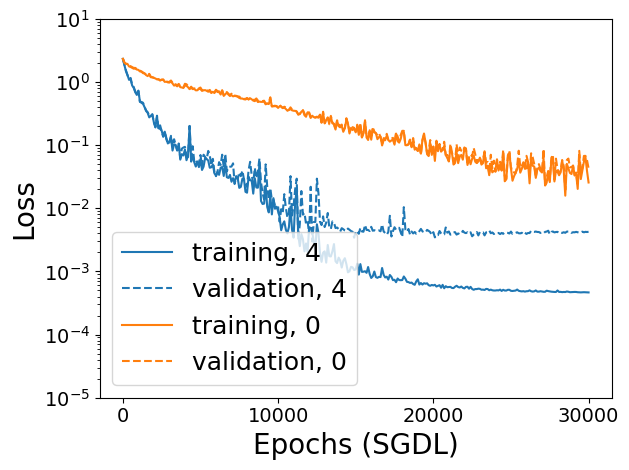}
    \end{subfigure}
   \begin{subfigure}{0.24\linewidth}
        \includegraphics[width=\linewidth]{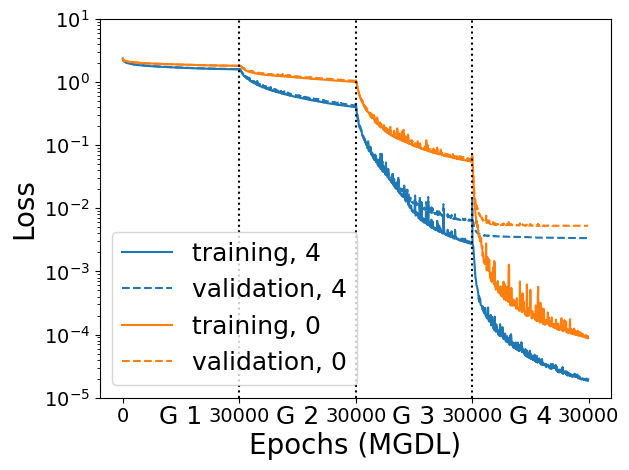}
   \end{subfigure}

	\caption{Comparison of the training and validation loss for SGDL (1st and 3rd subfigures) and MGDL (2nd and 4th subfigures) in settings 1 (1st and 2nd subfigures) and 2 (3rd and 4th subfigures). 
}
	\label{fig: compare MDNN and SDNN setting loss}
\end{figure}

{\bf 3.3 Regression on two-dimensional colored Images.}
In the third experiment, we compare performance of MGDL and SGDL for regression of two-dimensional color images.

We test the models with the `cat' image from website \href{https://live.staticflickr.com/7492/15677707699_d9d67acf9d_b.jpg}{Cat}, and the `sea' and `building' images from the Div2K dataset \cite{Agustsson2017}. The input to the models is the pixel coordinates, and the output is the corresponding RGB values. The training dataset consists of a regularly spaced grid containing $1/4$ of the image pixels, while the test dataset contains the full image pixels. We use peak signal-to-noise ratio (PSNR) defined as in \eqref{PSNR} to evaluate the accuracy of images obtained from MGDL and SGDL. PSNR values computed over the train and test dataset are denoted by TrPSNR and TePSNR, respectively.

\begin{figure}
  \centering
    \centering
   \begin{subfigure}{0.28\linewidth}    \includegraphics[width=\linewidth]{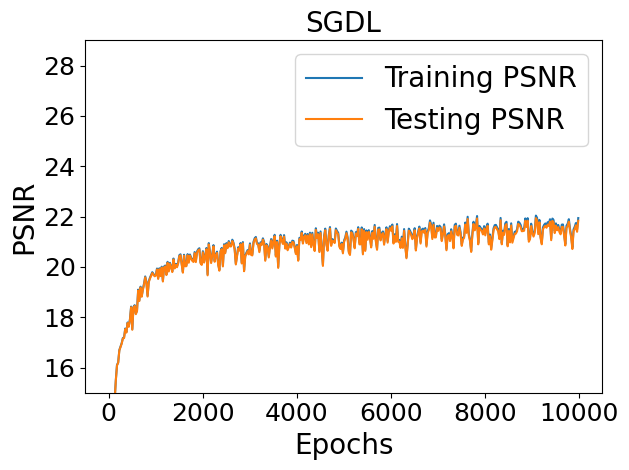}
    \caption{Cat: SGDL}
    \end{subfigure}
   \begin{subfigure}{0.28\linewidth}    \includegraphics[width=\linewidth]{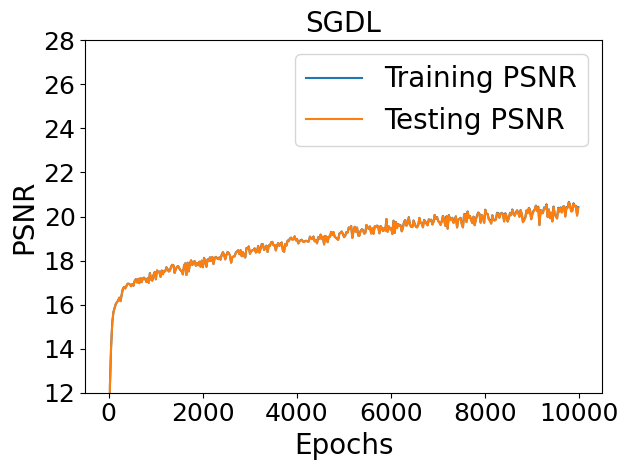}
    \caption{Sea: SGDL}
    \end{subfigure}
   \begin{subfigure}{0.28\linewidth}    \includegraphics[width=\linewidth]{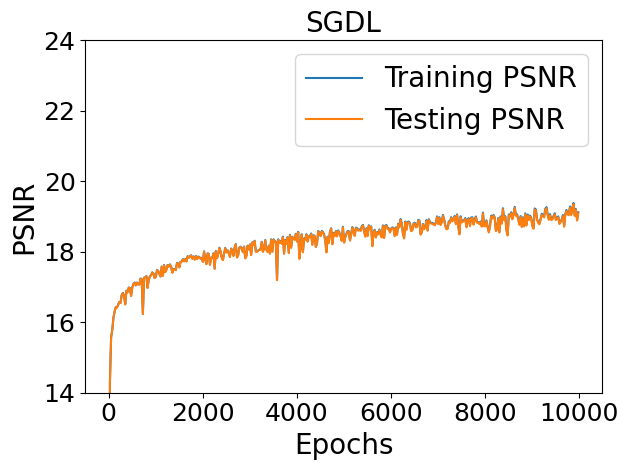}
    \caption{Building: SGDL}
    \end{subfigure}
    
   \begin{subfigure}{0.28\linewidth}    \includegraphics[width=\linewidth]{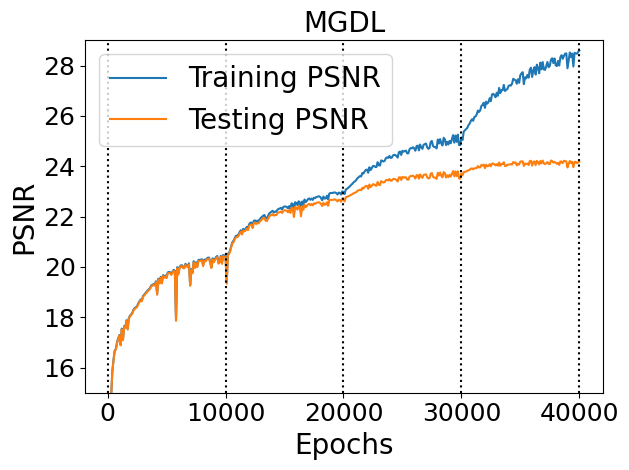}
    \caption{Cat: MGDL}
    \end{subfigure} 
   \begin{subfigure}{0.28\linewidth}    \includegraphics[width=\linewidth]{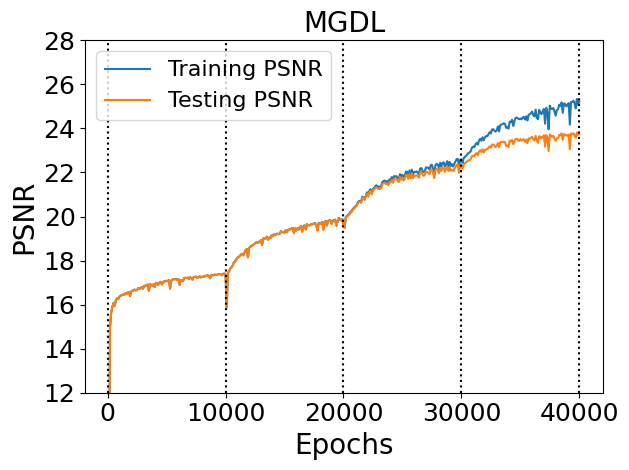}
    \caption{Sea: MGDL}
    \end{subfigure} 
   \begin{subfigure}{0.28\linewidth}    \includegraphics[width=\linewidth]{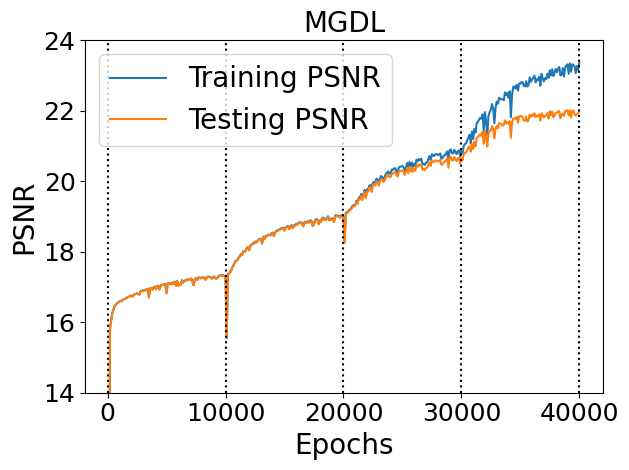}
    \caption{Building: MGDL}
    \end{subfigure} 

\caption{
Comparison of PSNR values for SGDL and MGDL on images Cat, Sea, and Building: SGDL (a)-(c), MGDL (d)-(f).} 
\label{fig: 2d image regression compare PSNR}
\end{figure}

Numerical results for this experiment are presented in Table \ref{tab: 2d image regression}, Figure \ref{fig: 2d image regression compare PSNR}, and Figures \ref{fig: 2d image regression cat}-\ref{fig: 2d image regression building} (in Appendix \ref{support material for example 3}). Table \ref{tab: 2d image regression} compares the PSNR values of images obtained from MGDL and SGDL. For MGDL, it is evident that adding more grades consistently improves the image quality, as both training and testing PSNR values increase substantially with the addition of each grade across all images Cat, Sea, and Building. MGDL outperforms SGDL by $2.35 \sim 3.93$ dB for the testing PSNR values. Specifically, for images Cat, Sea and Building, MGDL surpasses SGDL by 2.35,  3.93 and 2.88  dB, respectively. This demonstrates the superiority of MGDL in representing images compared to SGDL. Figure \ref{fig: 2d image regression compare PSNR} illustrates the training and testing PSNR values during the training process for the three images. It is evident that MGDL facilitates a smoother learning process as more grades are added, in comparison to SGDL. This observation aligns with the results presented in Table \ref{tab: 2d image regression}.

Predictions of each grade of MGDL and of SGDL are illustrated in Figures \ref{fig: 2d image regression cat}, \ref{fig: 2d image regression sea}, and \ref{fig: 2d image regression building} for images Cat, Sea, and Building, respectively. In all cases, grade 1 captures only the rough outlines of the images, representing the lower-frequency components. As we progress to grades 2, 3, and 4, more details corresponding to the higher-frequency components are added. This demonstrates the effectiveness of MGDL in progressively learning high-frequency features within the images. Moreover, the image quality achieved with MGDL is notably superior to that with SGDL.

\begin{small}
\begin{table}[]
    \centering
    \caption{PSNR comparison: SGDL versus MGDL.}
    \footnotesize
    \begin{tabular}{c||c|ccccc}
    \hline
    image & method & grade & learning rate & time (s) & TrPSNR   &TePSNR\\
\hline
\multirow{5}{*}{ cat  } &\multirow{4}{*}{ MGDL  }  & $1$ & $10^{-3}$ & $38$ & $20.45$ & $20.41$
\\
& & $2$  &  $10^{-3}$ & $40$ & $22.97$ & $22.67$\\
& & $3$  &  $5 \times 10^{-4}$ & $40$ & $25.14$ & $23.71$\\
& & $4$ &  $5 \times 10^{-4}$ & $28$ & $28.59$ & $24.18$\\
\cdashline{2-7}[1pt/1pt]
 & SGDL & & $5 \times 10^{-3}$ & $77$ & $21.93$ & $21.83$\\
 \hline
\multirow{5}{*}{ sea  } &\multirow{4}{*}{ MGDL  }  & $1$ & $10^{-2}$ & $160$ & $18.62$ & $18.62$
\\
& & $2$  &  $10^{-3}$ & $173$ & $21.57$ & $21.50$\\
& & $3$  &  $10^{-3}$ & $174$ & $24.17$ & $23.42$\\
& & $4$ &  $10^{-3}$ & $182$ & $27.31$ & $24.32$\\
\cdashline{2-7}[1pt/1pt]
 & SGDL & & $ 10^{-3}$ & $685$ & $20.43$ & $20.39$\\
 \hline
\multirow{5}{*}{ building   } &\multirow{4}{*}{ MGDL  }  & $1$ & $5\times 10^{-3}$ & $171$ & $17.30$ & $17.29$
\\
& & $2$  &  $5\times 10^{-3}$ & $181$ & $18.97$ & $18.95$\\
& & $3$  &  $10^{-3}$ & $182$ & $20.86$ & $20.67$\\
& & $4$ &  $10^{-3}$ & $182$ & $23.36$ & $21.97$\\
\cdashline{2-7}[1pt/1pt]
 & SGDL & & $10^{-3}$ & $742$ & $19.12$ & $19.09$\\
 \hline
\end{tabular}
    \label{tab: 2d image regression}
\end{table}
\end{small}

{\bf 3.4 Regression on MNIST data.}
In our fourth experiment, we apply MGDL to regress on high-dimension data, the MNIST data. We compare the training and validation loss, and TeRSE for SGDL and MGDL, when learning high-frequency features from the data. 

We set up this experiment following \cite{Rahaman2019}, with a focus on comparing performance of SGDL and MGDL in learning high-frequency features.  We choose the handwriting digits from MNIST dataset \cite{LeCun1998}, composed of 60,000 training samples and 10,000 testing samples of the digits ``0'' through ``9''. We represent each digit $j=0, 1, \ldots, 9$ by a one-hot vector $e_{j+1} \in \bR^{10}$, whose $(j+1)$-th component is one and all others are zero, and denote by $\tau_0:\bR^{784} \to \bR^{10}$ the classifier, which is a piecewise constant function defined by $\tau_0(\mathbf{x}):=e_{j+1}$ if $\mathbf{x}$ represents the digit $j$.
We split the available training samples to form the training and validation data, denoted as $\mathbb{D}_0:=\left\{\mathbf{x}_{\ell}, \tau_0(\mathbf{x}_{\ell})\right\}_{\ell \in \mathbb{N}_{n_{train}}}$ and $\mathbb{D}{'}_{0}:= \left\{\mathbf{x}{'}_{\ell}, \tau_0(\mathbf{x}{'}_{\ell})\right\}_{\ell \in \mathbb{N}_{n_{val}}}$ respectively, with $n_{train}:= 45,000$ and $n_{val} := 15,000$, and use the testing samples as the testing data, denoted as $\mathbb{D}{''}_0:=\left\{\mathbf{x}{''}_{\ell}, \tau_0(\mathbf{x}{''}_{\ell})\right\}_{\ell \in \mathbb{N}_{n_{test}}}$ with $n_{test}:=10,000$. Clearly, $\mathbb{D}_0$,  $\mathbb{D}_0{'}$ and $\mathbb{D}{''}$ are subsets of $\{[0, 1]^{784}, \left\{e_{j+1}\right\}_{j=0}^{9}\}$. Letting 
$
\psi_{\kappa}(\mathbf{x}) := \sin(2\pi\kappa\norm{\mathbf{x}}_2),
$
corresponding to a radial wave defined on the input space  $\bR^{784}$, we define the target function by
$
\tau_{\beta, \kappa}(\mathbf{x}) := \tau_0(\mathbf{x})\left( 1+\beta \psi_{\kappa}  (\mathbf{x})\right),  
$ 
where $\kappa$ is the frequency of the wave and $\beta$ is the amplitude. Note that $\tau_0$ and $\beta \psi_{\kappa}$ contribute respectively the lower-frequency and high-frequency components (regarded as noise) of the target function $\tau_{\beta, \kappa}$, as discussed in \cite{Rahaman2019}. The modified training and validation data denoted by $\mathbb{D}_{\beta, \kappa}:=\left\{\mathbf{x}_{\ell}, \tau_{\beta, \kappa}(\mathbf{x}_{\ell})\right\}_{\ell \in \mathbb{N}_{n_{train}}}$ and $\mathbb{D}'_{\beta, \kappa}:=\left\{\mathbf{x}'_{\ell}, \tau_{\beta, \kappa}(\mathbf{x}'_{\ell})\right\}_{\ell \in \mathbb{N}_{n_{val}}}$, respectively, are used to train DNNs. Our goal is to use SGDL and MDGL to regress the modified data $\mathbb{D}_{\beta, \kappa}$ through minimizing their respective training loss, to compare their robustness to noise. The training loss is evaluated on  $\mathbb{D}_{\beta, \kappa}$ and validation loss is on $\mathbb{D}'_{\beta, \kappa}$. TrRSE, VaRSE, and TeRSE are evaluated on $\mathbb{D}_{\beta, \kappa}$ $\mathbb{D}'_{\beta, \kappa}$, and $\mathbb{D}''_{0}$, respectively, noting that $\mathbb{D}''_{0}$ are test data without noise.

We choose the amplitude $\beta$ from $\left\{0.5, 1, 3, 5\right\}$. For each $\beta$, we vary the frequency $\kappa$ from $\left\{1, 3, 5, 7, 10, 50\right\}$. These choices of $\beta$ and $\kappa$ result in functions $\tau_{\beta,\kappa}$ that have higher frequencies than those studied in  \cite{Rahaman2019}. 
Figures \ref{fig: MINIST combined SDNN MDNN vLOSS} and \ref{fig: MINIST combined SDNN MDNN vLOSS REDO} (in Appendix \ref{Experimental details 3.4})
compare the training and validation loss versus training time of SGDL (structure \eqref{single equal structure MNIST}) and MGDL (structures \eqref{MNIST-structure1} and \eqref{MNIST-structure2} respectively) for different values of frequency $\kappa$ and amplitude $\beta$ (depicted in the figure). We observe that for small $\beta$, for example, $\beta = 0.5$,  the results of the two models are comparable. When $\beta=1$, the training loss for SGDL keeps decreasing as training time increases, while the validation loss initially decreases and starts to increase after training of 1,600 seconds, and unlike SGDL, both the training loss and validation loss for MGDL keep decreasing. This trend indicates that over-fitting phenomenon occurs for SGDL and suggests that MGDL has a superior generalization capability. Further increasing the value of $\beta$, for example, $\beta=3$ and $\beta=5$, the over-fitting phenomenon occurs in both of the models. An increase in $\beta$ corresponds to a higher proportion of high-frequency components within $\tau_{\beta, \kappa}$. This increased proportion of the high-frequency component significantly impacts the validation loss of SGDL, a trend also observed in MGDL. However, the effect is comparatively less with MGDL than with SGDL. We present in Table \ref{tab: MNIST} TrRSE, VaRSE, and TeRSE of SGDL (structure \eqref{single equal structure MNIST}) and MGDL (structure \eqref{MNIST-structure1}) for $\beta := 1$, where the results are obtained by choosing $t_{min}:=10^{-5}$ and $t_{max}:=10^{-4}$, and the batch size to be `Full',  for all cases of $\kappa$ and for both SGDL and MGDL. We observe that while TrRSE of SGDL is smaller than that of MGDL, both VaRSE, and TeRSE of SGDL are larger than the corresponding values of MGDL, suggesting occurrence of  overfitting with SGDL. MGDL's improvement in accuracy, measured by TeRSE, is about $27 \sim 29\%$. This experiment reveals that MGDL is a promising model for learning higher-dimensional data with a higher proportion of high-frequency features.

\begin{small}

\begin{table}[]
    \centering
    \caption{Accuracy comparison: SGDL versus MGDL with $\beta = 1$. 
    }
    \begin{tabular}{c|c|rccc}
    \hline
    $\kappa$ & method & time (s) & TrRSE  &VaRSE  & TeRSE\\
\hline
 \multirow{ 2}{*}{ 1 } & SGDL& $3,298$& $\mathbf{3.1\times 10^{-1}}$ & $4.1 \times 10^{-1}$  & $1.1 \times 10^{-1}$
\\
 & MGDL  &  $3,109$ & $3.5 \times 10^{-1}$ & $\mathbf{3.9\times 10^{-1}}$ & $\mathbf{8.0 \times 10^{-2}}$
\\
 \cdashline{1-6}[1pt/1pt]
\multirow{ 2}{*}{ 5 } &SGDL & $3,333$ & $\mathbf{3.0\times 10^{-1}}$ & $4.1\times 10^{-1}$  & $1.1 \times 10^{-1}$
\\
 & MGDL  &  $3,461$ & $3.5 \times 10^{-1}$ & $ \mathbf{3.9 \times 10^{-1}}$  & $\mathbf{7.8 \times 10^{-2}}$
\\
 \cdashline{1-6}[1pt/1pt]
 \multirow{ 2}{*}{ 10 } & SGDL  & $3,199$ & $\mathbf{3.1\times 10^{-1}}$ & $4.1\times 10^{-1}$  & $1.0 \times 10^{-1}$
\\
 & MGDL &  $3,448$ & $3.5 \times 10^{-1}$ & $\mathbf{3.9 \times 10^{-1}}$  & $\mathbf{7.8 \times 10^{-2}}$
\\
 \cdashline{1-6}[1pt/1pt]
 \multirow{ 2}{*}{ 50 }  & SGDL & $3,168$ & $\mathbf{3.0\times 10^{-1}}$ & $4.1\times 10^{-1}$ & $1.1 \times 10^{-1}$
\\
 & MGDL & $3,484$ &$ 3.5\times 10^{-1}$ & $\mathbf{3.9 \times 10^{-1}}$  & $\mathbf{7.9 \times 10^{-2}}$
\\
\hline
    \end{tabular}
    \label{tab: MNIST}
\end{table}
\end{small}

\vspace{-3mm}
\section{Conclusion}
\vspace{-3mm}
By observing that a high-frequency function may be well-represented by composition of several lower-frequency functions, we have proposed a novel approach to learn such a function. The proposed approach decomposes the function into a sum of multiple frequency components, each of which is compositions of lower-frequency functions. By leveraging the MGDL model, we can express high-frequency components by compositions of multiple SNNs of low-frequency. We have conducted numerical studies of the proposed approach by applying it to one-dimensional synthetic data, two-dimensional manifold data, colored images, and higher-dimensional MNIST data. Our studies have concluded that MGDL can effectively learn high-frequency components within data.
Moreover, the proposed approach is easy to implement and not limited by dimension of the input variable. Consequently, it offers a promising approach to learn high-frequency features within (high-dimensional) dataset.


\noindent
\textbf{Limitation}: Mathematical understanding of the spectral bias of DNNs is absent. Theoretical foundation for MGDL to address the spectral bias issue needs to be established. Numerical studies are preliminary, limited to four examples. More extensive numerical experiments will be conducted in our future work.


\noindent
\textbf{Acknowledgments:} Y. Xu is indebted to Professor Wei Cai of Southern Methodist University for insightful discussion of the spectral bias of DNNs.
Y. Xu is supported in part by the US National Science Foundation under grant DMS-2208386, and by the US National Institutes of Health under grant R21CA263876.


\appendix

\section{Analysis of the Multi-Grade Deep Learning Model}\label{AnalysisofMGDL}

It was established in \cite{XuMultigrade2023first} that when the loss function is defined in terms of the mean square error, MGDL either learns the zero function or results in a strictly decreasing residual error sequence in each grade.

\begin{theorem}\label{Theorem_1}
 Let  $\mathbb{D}$ be a compact subset of $\mathbb{R}^s$ and $L_2(\mathbb{D}, \mathbb{R}^t)$ denote the space of $t$-dimensional vector-valued square integral functions on $\mathbb{D}$. If $\mathbf{f} \in L_2(\mathbb{D}, \mathbb{R}^t)$,  
then for all $i=1, 2, \ldots$,
    $$
    \mathbf{f} = \sum_{l=1}^i \mathbf{f}_l +\mathbf{e}_i, \quad \mathbf{f}_l:=\mathcal{N}^*_{l}\circ\mathcal{N}^*_{l-1}\circ\cdots\circ\mathcal{N}^*_{1}.
    $$
    where $\mathcal{N}^*_{l}$ is the SNN learned in grade $l$ of MGDL, and
     for $i=1,2, \ldots$, either $\mathbf{f}_{i+1} = \mathbf{0}$ or 
    $$
    \left\|\mathbf{e}^*_{i+1} \right\| < \left\|\mathbf{e}^*_{i} \right\|.
    $$
\end{theorem}

All numerical examples presented in this paper validate Theorem \ref{Theorem_1}.

\section{Supporting material for Section 3}
\label{Experimental details}

We provide in this appendix details for the numerical experiments in Section \ref{numerical experiments}, including computational resources, network structures of SGDL and MGDL, the choice of activation function, as well as the optimizer and parameters used in the optimization process, and some supporting figures.

The experiments conducted in Sections 3.1, 3.2, and 3.4 of Section \ref{numerical experiments} were performed on X86\_64 server equipped with an Intel(R) Xeon(R) Gold 6148 CPU @ 2.4GHz (40 slots) or Intel(R) Xeon(R) CPU E5-2698 v3 @ 2.30GHz (32 slots). 
In contrast, the experiments described in Section 3.3 of Section \ref{numerical experiments} were performed on X86\_64 server equipped with AMD 7543 @ 2.8GHz (64 slots) and AVX512, 2 x Nvidia Ampere A100 GPU.


We choose ReLU as the activation function as in  \cite{Rahaman2019} for all the four experiments.
For each experiment, for SGDL we test several network structures and choose the one produces the best performance. We then design MGDL, having the same total number of layers and the same number of neurons for each layer as the chosen SGDL structure, but their parameters are trained in multiple grades as described in Section \ref{section: Frequency analysis}. Details of the network structure will be described for each experiment.




The optimization problems for both SGDL and MGDL
across the four experiments are solved by the Adam  method \cite{Kingma2014} with `Xavier' initialization \cite{Glorot2020}. In Sections 3.1, 3.2, and 3.4, the learning rate $t_k$ for the $k$-th epoch decays exponentially with each epoch \cite{Jiang2024}, calculated as
$t_k := t_{max} e^{-\gamma k}$,    
where $\gamma := (1/K)\ln(t_{max}/t_{min})$ represents the decay rate, with $K$ being the total number of training epochs, $t_{max}$ and $t_{min}$ denoting the predefined maximum and minimum learning rates, respectively. In Section 3.3, we employed a fixed learning rate for both SGDL and MGDL, as numerical results indicate that the exponential decay learning rate performs poorly. 
We observe that when a network structure of SGDL is split into multiple grades of MGDL, the combined computing time required to train all grades in MGDL, each for $K$ epochs, is comparable to the computing time needed to train the SGDL with $K$ epochs, due to only SNNs involved in MGDL. Therefore, in all the four experiments, we train SGDL and all grades of MGDL for the same $K$ epochs. We test SGDL and MGDL with the same set of the algorithm parameters, including $t_{min}, t_{max}$, batch size, and $K$, for all the experiments. Optimal parameters were selected based on the lowest VaRSE value for Sections 3.1 and 3.2, the highest PSNR value for Section 3.3, and the lowest validation loss for Section 3.4, within the range of parameters to be described for each example.

\subsection{Section 3.1}\label{Experimental details 3.1}

For settings 1, 2, and 3, the network structure for SGDL is
\begin{equation}\label{synthetic single increase structure}
[1] \to [256]\times 8 \to  [1],    
\end{equation}
where $[n]\times N$ indicates $N$ hidden layers, each with $n$ neurons. 
The network structure for each grade of MGDL is given by
\begin{align*}
    &\text{Grade 1:}\quad [1] \to [256]  \times 2 \to [1]\\
    &\text{Grade 2:}\quad [1] \to [256]_F \times 2 \to [256] \times 2 \to [1]\\
    &\text{Grade 3:}\quad [1] \to [256]_F\times 4  \to  [256] \times 2 \to [1]\\
    &\text{Grade 4:}\quad [1] \to [256]_F\times 6   \to [256] \times 2 \to [1].
\end{align*}
Here, $[n]_F$ indicates a layer having $n$ neurons with parameters, trained in the previous grades, remaining fixed during the training of the current grade. We employ this notation across the numerical experiment section without further mentioning. 
For setting 4, the most challenging case, we employ a deeper structure for SGDL, with two more hidden layers in addition to \eqref{synthetic single increase structure}. Correspondingly,  for MGDL we add one more grade:
\begin{align*}
    &\text{Grade 5:}\quad [1] \to [256]_F\times 8   \to [256] \times 2 \to [1].
\end{align*}

We now describe the search range of the parameters for all the four settings. 
We let $I_1 := \left\{10^{-4}, 10^{-5}, 10^{-6}\right\}$, $I_2 := \left\{10^{-3}, 5 \times 10^{-4}, 1\times 10^{-4}\right\}$, $I_3:=\left\{ 10^{-5}, 10^{-6}, 10^{-7}\right\}$ and $I_4:=\left\{5\times 10^{-5}, 10^{-5}\right\}$.
Then for both SGDL and MGDL, we test the pair $(t_{min}, t_{max})$ from all possible cases in the set $ \left(I_1 \times I_2\right) \cup \left(I_3 \times I_4\right)$.
The batch size is chosen from $256, 512$ or the full gradient (denoted by `Full') for each epoch. The total epoch number $K$ is set to be 30,000. 

\begin{figure}
  \centering
   \begin{subfigure}{0.24\linewidth}  \includegraphics[width=\linewidth]{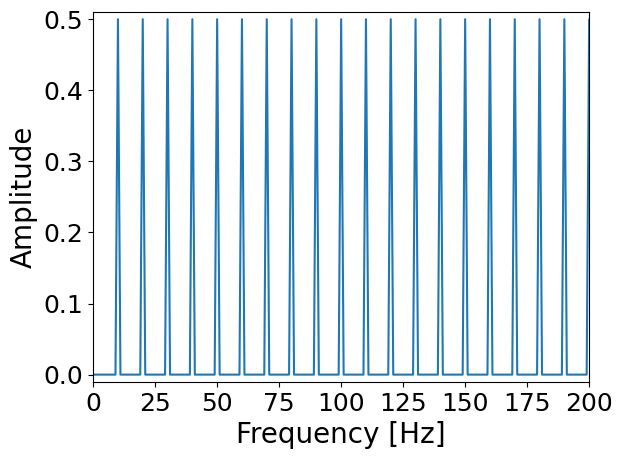}
   \end{subfigure}
   \begin{subfigure}{0.24\linewidth}  \includegraphics[width=\linewidth]{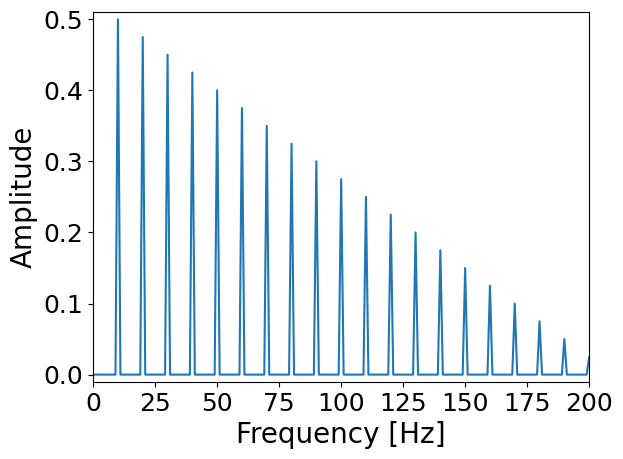}
   \end{subfigure}
   \begin{subfigure}{0.24\linewidth}
        \includegraphics[width=\linewidth]{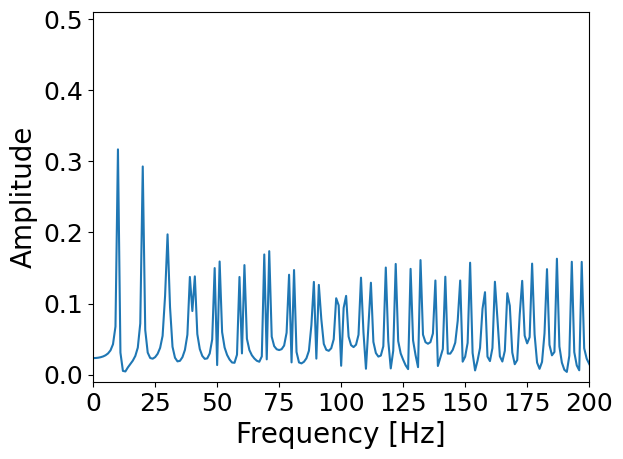}
    \end{subfigure}
\begin{subfigure}{0.24\linewidth}
        \includegraphics[width=\linewidth]{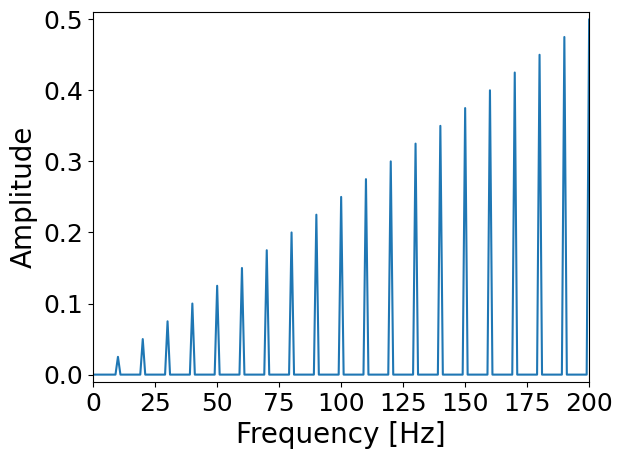}
    \end{subfigure}
	\caption{
 Amplitude versus one-side frequency plots for $\lambda$ across settings 1-4.}
	\label{fig: frequency constant increase vary original function}
\end{figure}

\begin{figure}
  \centering
      \begin{minipage}{1.0\textwidth}
        \centering
  \begin{subfigure}{0.24\linewidth}
    \includegraphics[width=\linewidth]{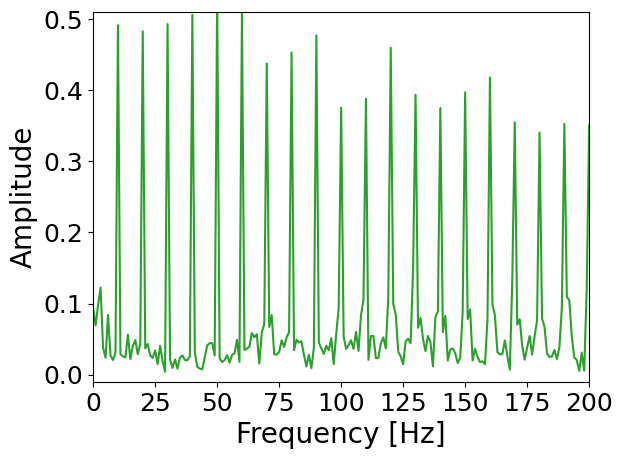}
  \end{subfigure}
  \begin{subfigure}{0.24\linewidth}
    \includegraphics[width=\linewidth]{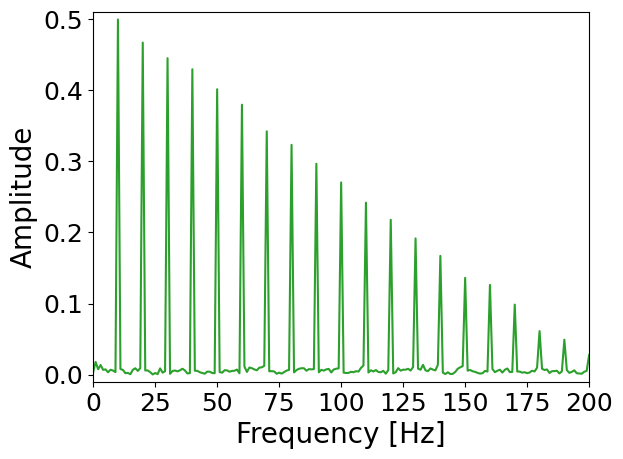}
  \end{subfigure}
  \begin{subfigure}{0.24\linewidth}
    \includegraphics[width=\linewidth]{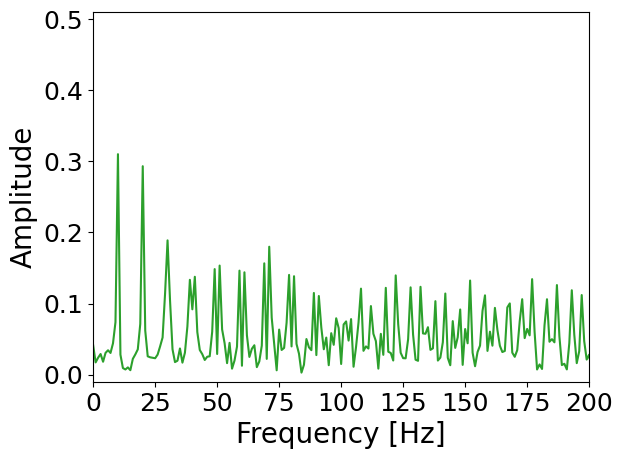}
  \end{subfigure}
  \begin{subfigure}{0.24\linewidth}
    \includegraphics[width=\linewidth]{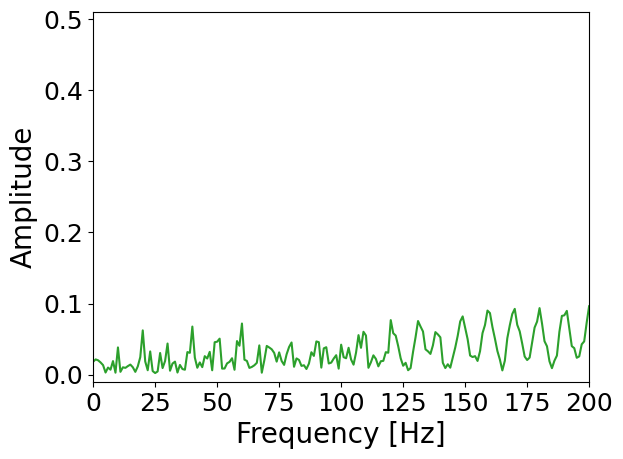}
  \end{subfigure}
  
  \begin{subfigure}{0.24\linewidth}
    \includegraphics[width=\linewidth]{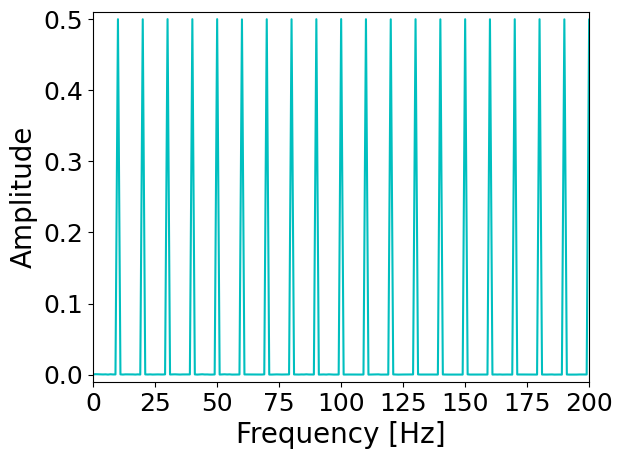}
  \end{subfigure}
  \begin{subfigure}{0.24\linewidth}
    \includegraphics[width=\linewidth]{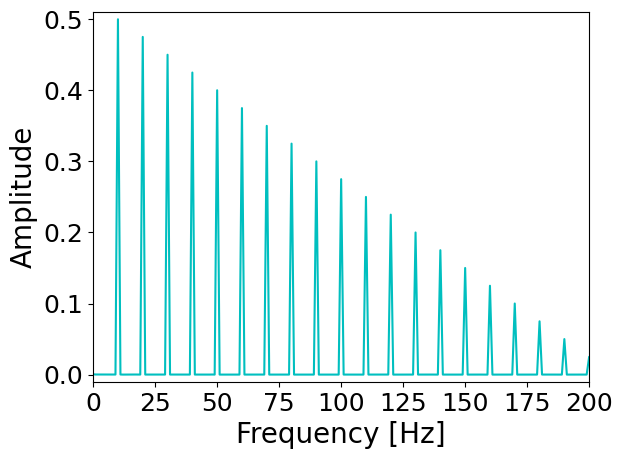}
  \end{subfigure}
  \begin{subfigure}{0.24\linewidth}
    \includegraphics[width=\linewidth]{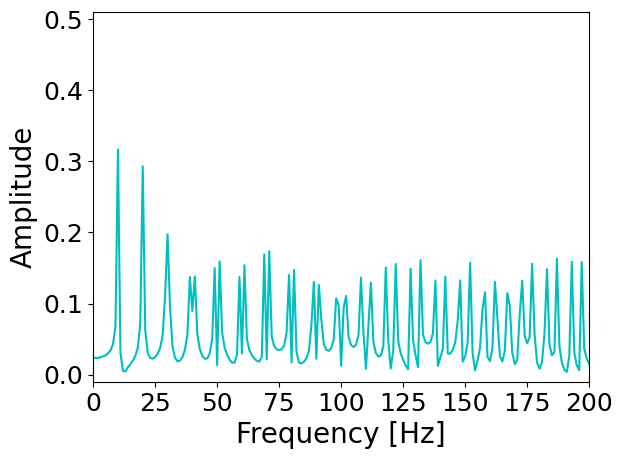}
  \end{subfigure}
  \begin{subfigure}{0.24\linewidth}
    \includegraphics[width=\linewidth]{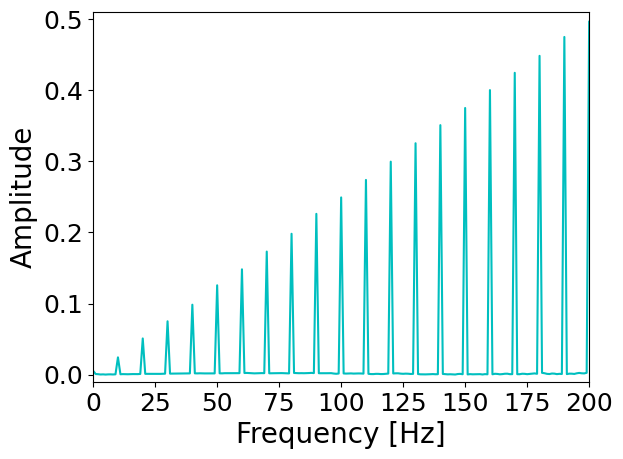}
  \end{subfigure}
\end{minipage}
  \caption{
    Comparison of SGDL (1st row) and MGDL (2nd row): Amplitude 
    across settings 1-4. 
  }
  \label{fig: frequency constant increase vary}
\end{figure}

\begin{figure}
    \centering
    \begin{minipage}{1.0\textwidth}
        \centering
        \begin{subfigure}{0.24\textwidth}
            \includegraphics[width=\linewidth]{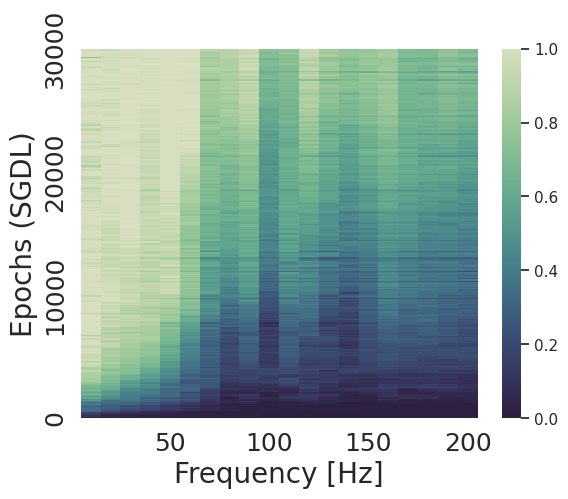}
        \end{subfigure}
        \begin{subfigure}{0.24\textwidth}
            \includegraphics[width=\linewidth]{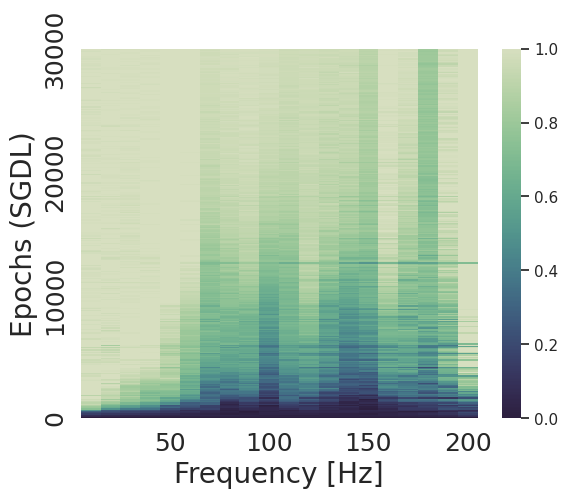}
        \end{subfigure}
        \begin{subfigure}{0.24\textwidth}
            \includegraphics[width=\linewidth]{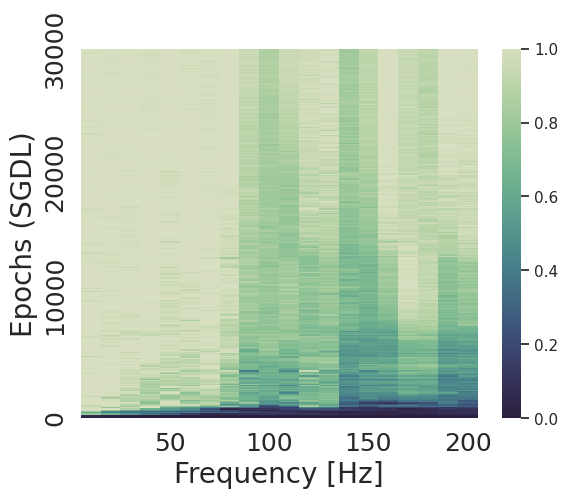}
        \end{subfigure}
        \begin{subfigure}{0.24\textwidth}
            \includegraphics[width=\linewidth]{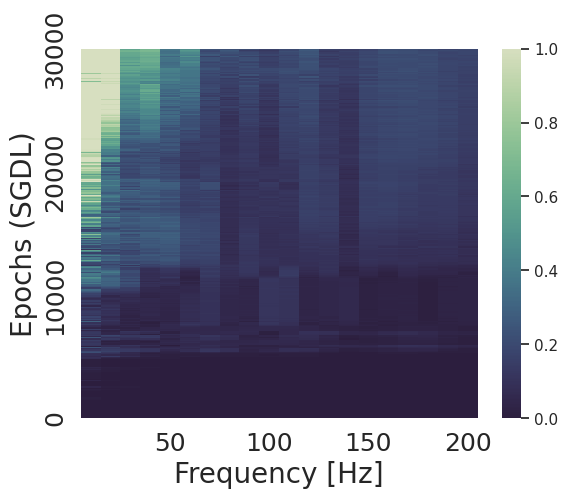}
        \end{subfigure}
        \begin{subfigure}{0.24\textwidth}
            \includegraphics[width=\linewidth]{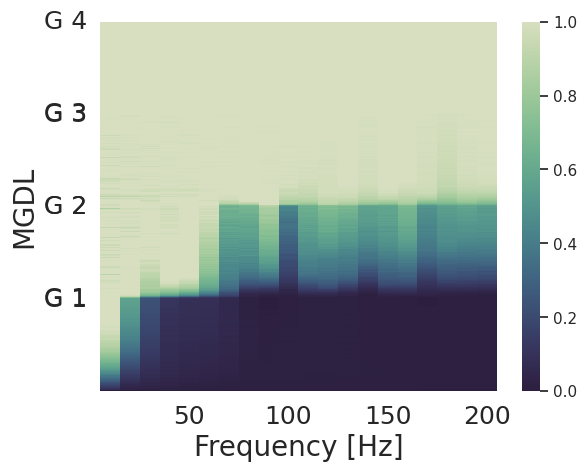}
        \end{subfigure}
        \begin{subfigure}{0.24\textwidth}
            \includegraphics[width=\linewidth]{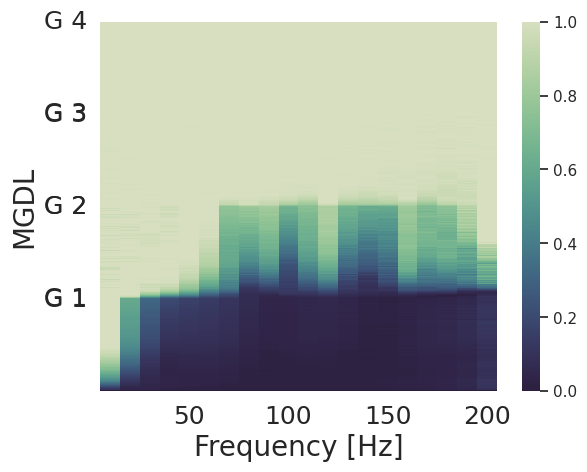}
        \end{subfigure}
        \begin{subfigure}{0.24\textwidth}
            \includegraphics[width=\linewidth]{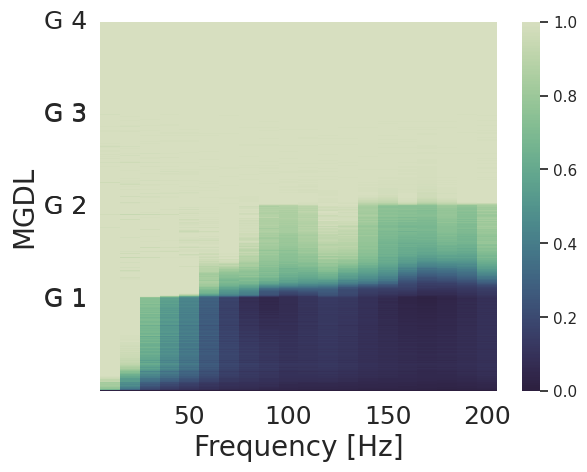}
        \end{subfigure}
        \begin{subfigure}{0.24\textwidth}
            \includegraphics[width=\linewidth]{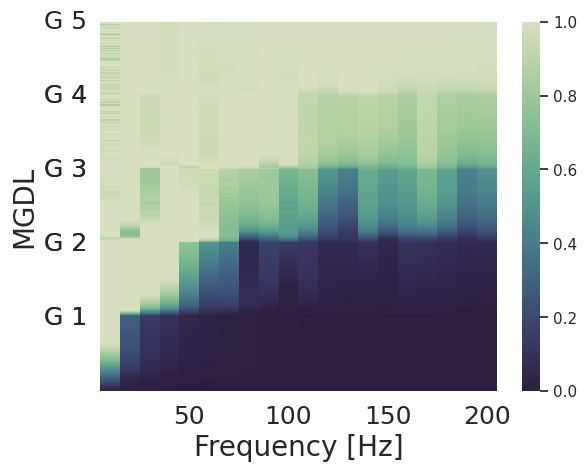}
        \end{subfigure}
    \end{minipage}
    \caption{Comparison of SGDL (1st row) and MGDL (2nd row) for  settings 1-4 of Section 3.1: The evolution of spectrum.
      }
    \label{fig: spetral compare SDNN and MDNN}
\end{figure}

The supporting figures for this experiment include: 
\begin{itemize}
    \item Figure \ref{fig: frequency constant increase vary original function}, presenting amplitude versus one-side frequency plots for $\lambda$ across settings 1-4; 
    \item Figure \ref{fig: frequency constant increase vary}, comparing the amplitude of one-side frequency for SGDL and MGDL across settings 1-4; 
    \item Figure \ref{fig: spetral compare SDNN and MDNN}, illustrating the evolution of spectrum comparison between SGDL and MGDL for settings 1-4. 
\end{itemize}
In Figure \ref{fig: spetral compare SDNN and MDNN}, the colors in these subfigures show the measured amplitude of the network spectrum at the corresponding frequency, normalized by the amplitude of $\lambda$ at the same frequency. The colorbar is clipped between $0$ and $1$, indicating approximation accuracy from the worst to the best when changing from $0$ to $1$.

\subsection{Section 3.2}\label{Experimental details 3.2}

The network structure that we use for SGDL is
\begin{equation*}\label{manifold single increase structure}
[2] \to [256] \times 8 \to  [1], 
\end{equation*}
and the grade network structure for MGDL is 
\begin{align*}
    &\text{Grade 1:}\ \ [2] \to [256]  \times 2 \to [1]\\
    &\text{Grade 2:}\ \ [2] \to [256]_F \times 2 \to [256] \times 2 \to [1]\\
    &\text{Grade 3:}\ \ [2] \to [256]_F\times 4  \to  [256] \times 2 \to [1]\\
    &\text{Grade 4:}\ \ [2] \to [256]_F\times 6   \to [256] \times 2 \to [1].
\end{align*}

For choices of $t_{min}$ and $t_{max}$, we let $I_1:= \left\{10^{-4}, 10^{-5}, 10^{-6}\right\}$ and $I_2:=\left\{1\times 10^{-3}, 5\times 10^{-4}, 1\times 10^{-4}\right\}$. For both SGDL and MGDL, we test the pair $(t_{min}, t_{max})$ from all possible cases in the set $I_1 \times I_2$, the batch size is chosen from $512, 1024$, or the full gradient for each epoch, and the total number of the epochs $K$ is set to be 30,000.

\begin{figure}
  \centering
  \begin{minipage}{1.0\textwidth}
  \centering
   \begin{subfigure}{0.24\linewidth}        
   \includegraphics[width=\linewidth]{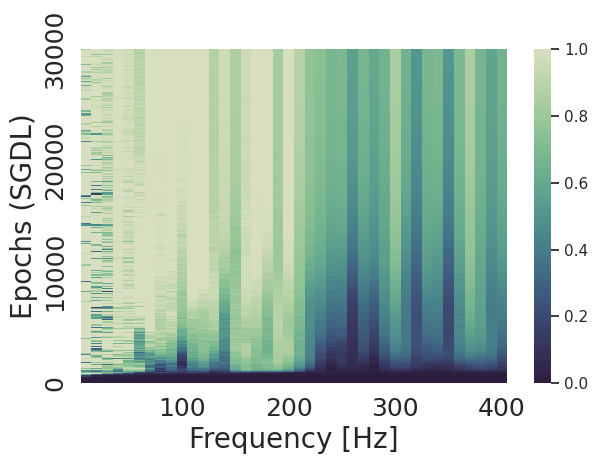}
    \end{subfigure}
  \begin{subfigure}{0.24\linewidth}
       \includegraphics[width=\linewidth]{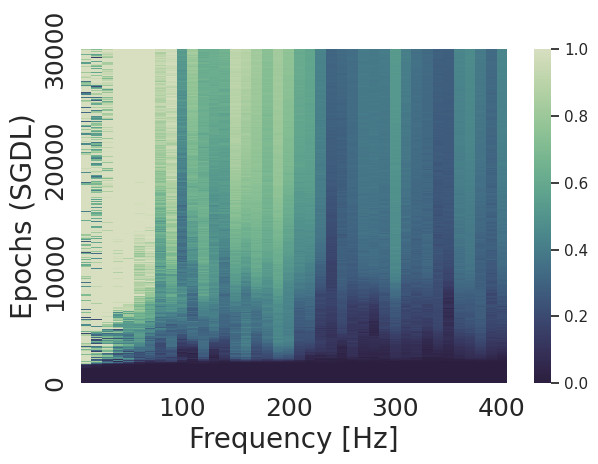}
   \end{subfigure}
   \begin{subfigure}{0.24\linewidth}
        \includegraphics[width=\linewidth]{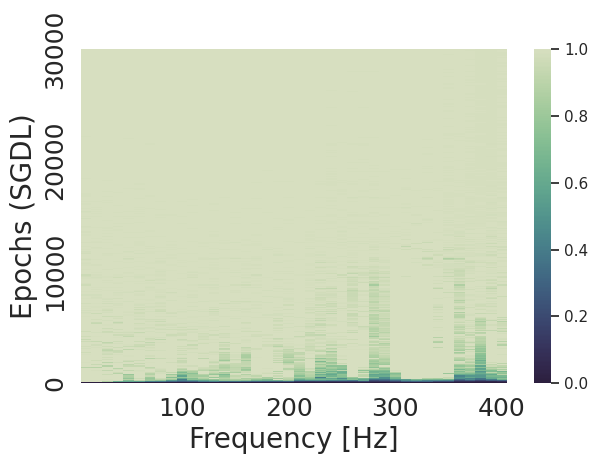}
    \end{subfigure}
  \begin{subfigure}{0.24\linewidth}
       \includegraphics[width=\linewidth]{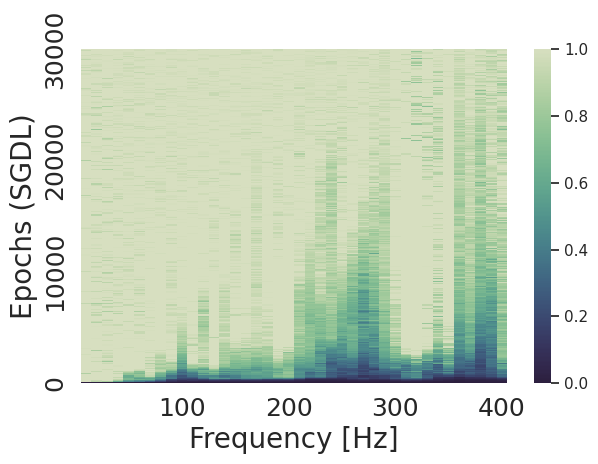}
   \end{subfigure}
   \begin{subfigure}{0.24\linewidth}
        \includegraphics[width=\linewidth]{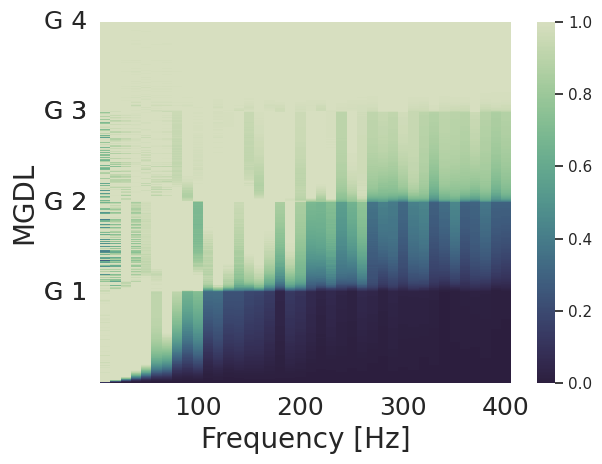}
    \end{subfigure}
    \begin{subfigure}{0.24\linewidth}
         \includegraphics[width=\linewidth]{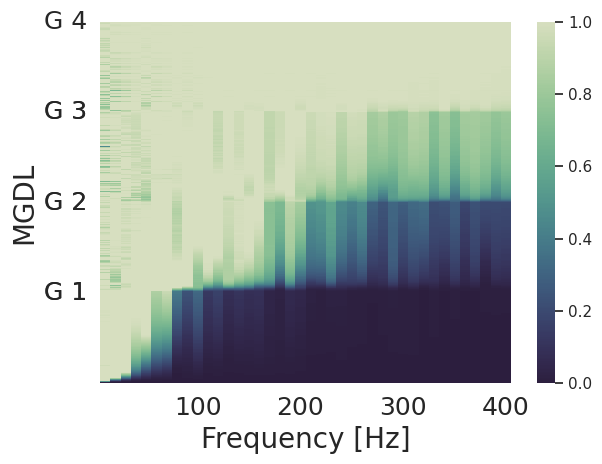}
    \end{subfigure}
   \begin{subfigure}{0.24\linewidth}
        \includegraphics[width=\linewidth]{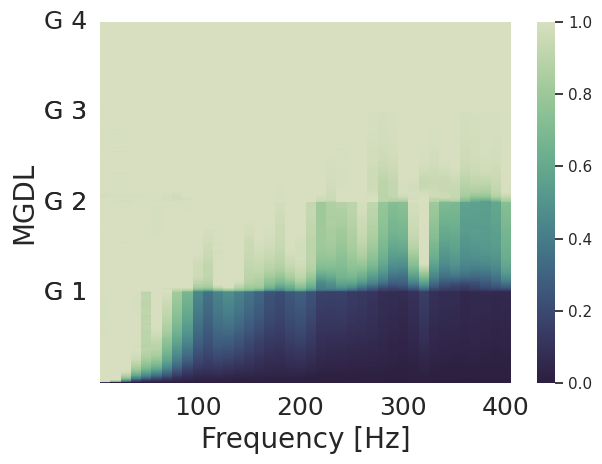}
    \end{subfigure}
    \begin{subfigure}{0.24\linewidth}
         \includegraphics[width=\linewidth]{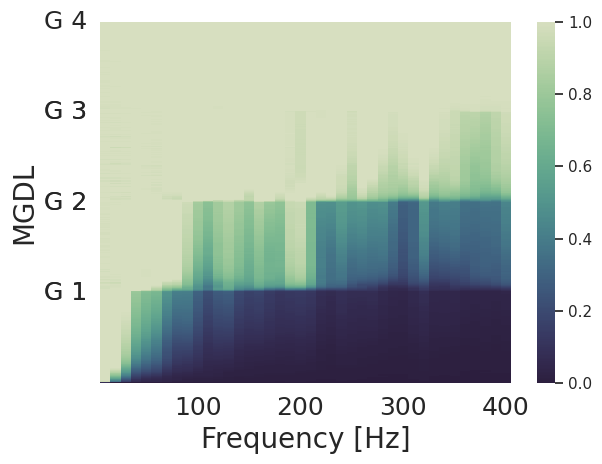}
    \end{subfigure}
   \end{minipage}
	\caption{Comparsion of SGDL (1st row) and MGDL (2nd row) for settings 1 and 2 of Section 3.2: The evolution of spectrum (the first and second columns for the learned functions on manifolds $\gamma_q$ with $q = 4$ and $q=0$, respectively, for setting 1, while the third and fourth columns for setting 2). 
}
	\label{fig: compare MDNN and SDNN setting 1}
\end{figure}

The supporting figures for this experiment include Figure \ref{fig: compare MDNN and SDNN setting 1}, illustrating the evolution of spectrum comparison between SGDL and MGDL for settings 1-2 of Section 3.2. The mean of colorbar in Figure \ref{fig: compare MDNN and SDNN setting 1} is consistent with that in Figure \ref{fig: spetral compare SDNN and MDNN}.


\subsection{Section 3.3}\label{support material for example 3}

The network structure for SGDL is 
$$
[2] \to [256] \times 12 \to [3]
$$
and that for MGDL is
\begin{align*}
&\text{Grade 1:}\ \ [2] \to [256] \times 3  \to [3]\\ & \text{Grade 2:}\ \ [2] \to [256]_F \times 3 \to [256] \times 3  \to [3]\\
    &\text{Grade 3:}\ \ [2] \to [256]_F \times 6 \to  [256] \times 3 \to [3]\\
    &\text{Grade 4:}\ \ [2] \to [256]_F \times 9 \to  [256] \times 3 \to [3] \\ 
\end{align*}

For SGDL and all grades of MGDL, we select the learning rate from the set $\{10^{-2}, 5\times 10^{-3}, 10^{-3}, 5 \times 10^{-4}, 10^{-4}\}$, choose the full gradient for each epoch, and set the total epoch number $K$ to be $10,000$.
%
The quality of the reconstructed image in Section 3.3 is evaluated by the peak signal-to-noise ratio (PSNR) defined by
\begin{equation}\label{PSNR}
\text { PSNR }:=10 \log _{10}\left(\frac{ n \times 255^2}{\left\| \mathbf{v}- \mathbf{\hat v}\right\|_{\text{F}}^2}\right)
\end{equation}
where $\mathbf{v}$ is the ground truth image, $\mathbf{\hat v}$ is reconstructed image, $n$ is the number of pixels in $\mathbf{v}$, and $\norm{\mathbf{\cdot}}_{\text{F}}$ denotes the Frobenius norm of a matrix.

\begin{figure}
  \centering
    \centering
   \begin{subfigure}{0.32\linewidth}\includegraphics[width=\linewidth]{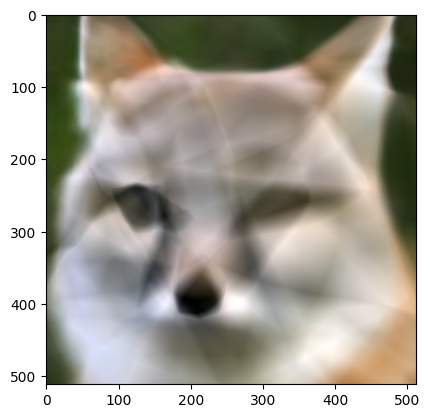}
    \caption{G 1: 20.41}
    \end{subfigure}  
   \begin{subfigure}{0.32\linewidth}  \includegraphics[width=\linewidth]{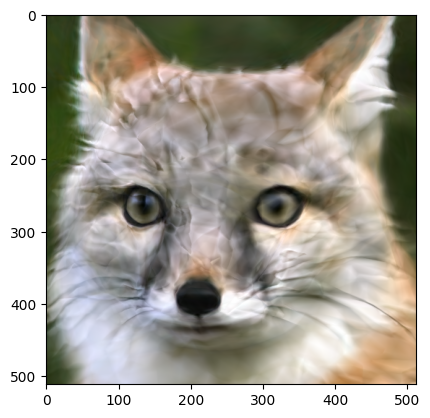}
   \caption{G 2: 22.67}
   \end{subfigure}
   \begin{subfigure}{0.32\linewidth}    \includegraphics[width=\linewidth]{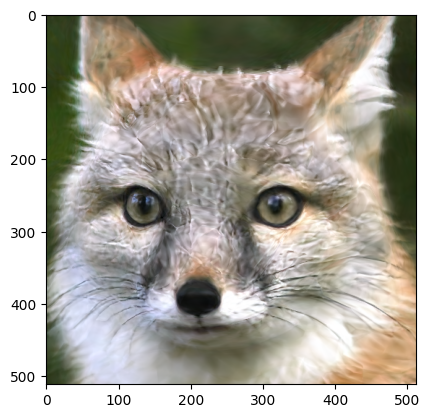}
    \caption{G 3: 23.71}
    \end{subfigure} 
    \begin{subfigure}{0.32\linewidth}  \includegraphics[width=\linewidth]{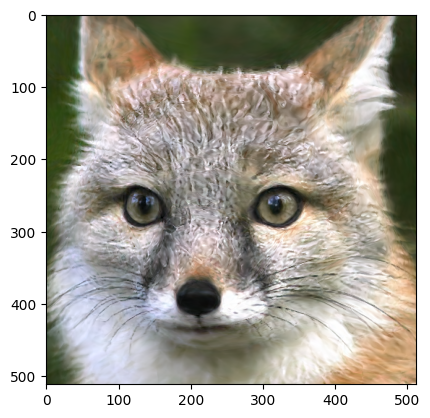}
   \caption{G 4: 24.18}
   \end{subfigure}
       \begin{subfigure}{0.32\linewidth}  \includegraphics[width=\linewidth]{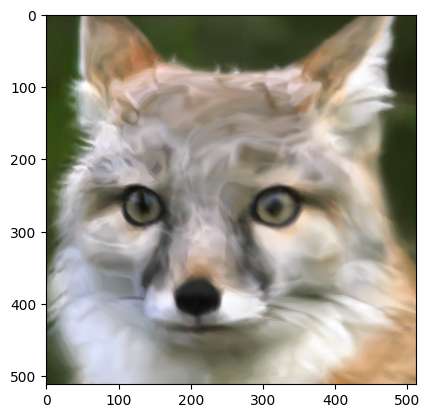}
   \caption{SGDL: 22.84}   
   \end{subfigure}
   \begin{subfigure}{0.32\linewidth}    \includegraphics[width=\linewidth]{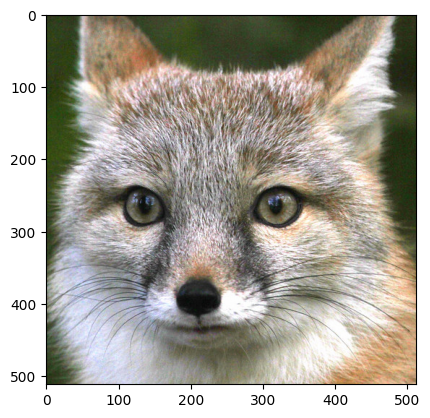}
    \caption{Ground Truth}
    \end{subfigure} 
\caption{
Comparison of MGDL and SGDL for image Cat. (a)-(d): Predictions of MGDL for grades 1-4, with the corresponding testing PSNR values indicated in the subtitles. (e): Prediction of SGDL with testing PSNR displayed in the subtitle. (f): Ground truth image} 
\label{fig: 2d image regression cat}
\end{figure}

\begin{figure}
  \centering
    \centering
   \begin{subfigure}{0.32\linewidth}\includegraphics[width=\linewidth]{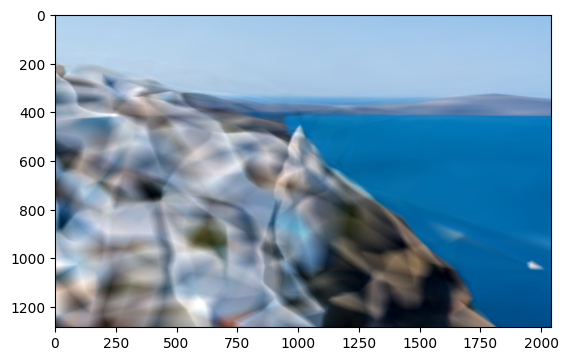}
    \caption{G 1: 18.62}
    \end{subfigure}  
   \begin{subfigure}{0.32\linewidth}  \includegraphics[width=\linewidth]{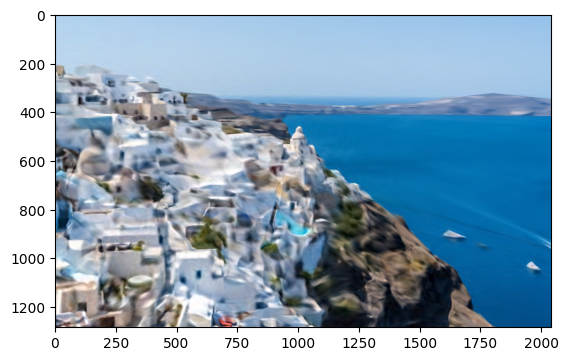}
   \caption{G 2: 21.50}
   \end{subfigure}
   \begin{subfigure}{0.32\linewidth}    \includegraphics[width=\linewidth]{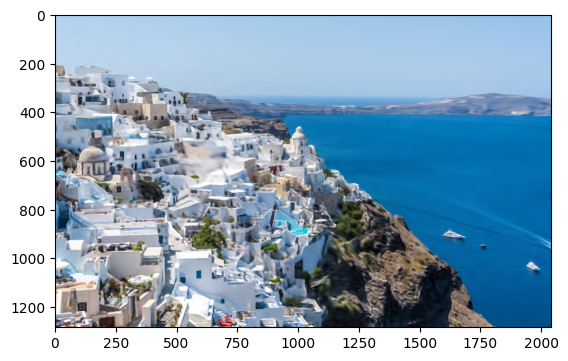}
    \caption{G 3: 23.42}
    \end{subfigure} 
    \begin{subfigure}{0.32\linewidth}  \includegraphics[width=\linewidth]{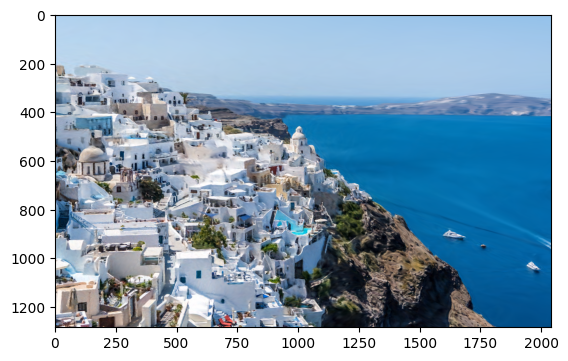}
   \caption{G 4: 24.32}
   \end{subfigure}
       \begin{subfigure}{0.32\linewidth}  \includegraphics[width=\linewidth]{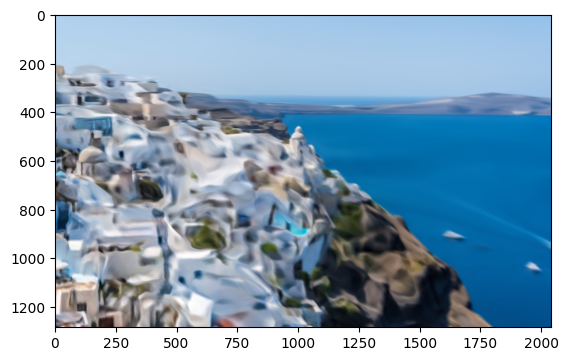}
   \caption{SGDL: 20.39}   
   \end{subfigure}
   \begin{subfigure}{0.32\linewidth}    \includegraphics[width=\linewidth]{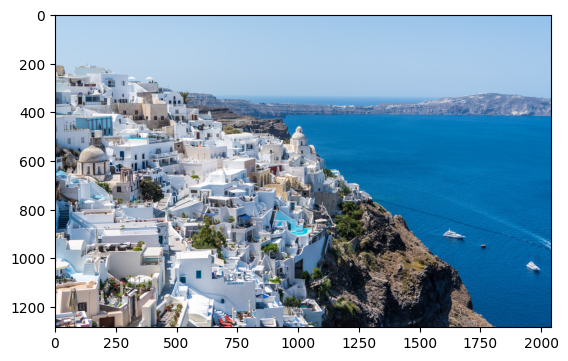}
    \caption{Ground Truth}
    \end{subfigure} 
\caption{
Comparison of MGDL and SGDL for image Sea. (a)-(d): Predictions of MGDL for grades 1-4, with the corresponding testing PSNR values indicated in the subtitles. (e): Prediction of SGDL with testing PSNR displayed in the subtitle. (f): Ground truth image} 
\label{fig: 2d image regression sea}
\end{figure}

\begin{figure}
  \centering
    \centering
   \begin{subfigure}{0.32\linewidth}\includegraphics[width=\linewidth]{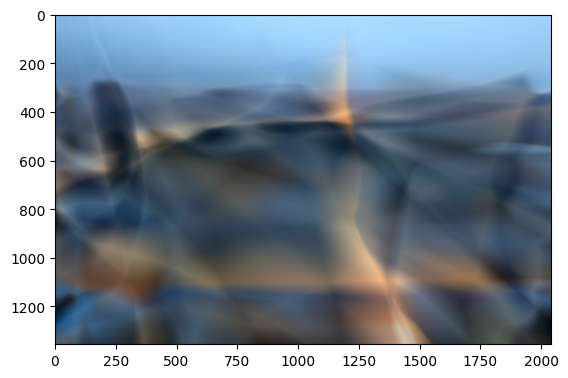}
    \caption{G 1: 17.29}
    \end{subfigure}  
   \begin{subfigure}{0.32\linewidth}  \includegraphics[width=\linewidth]{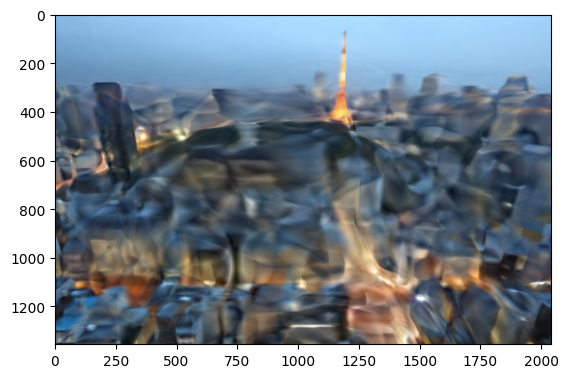}
   \caption{G 2: 18.95}
   \end{subfigure}
   \begin{subfigure}{0.32\linewidth}    \includegraphics[width=\linewidth]{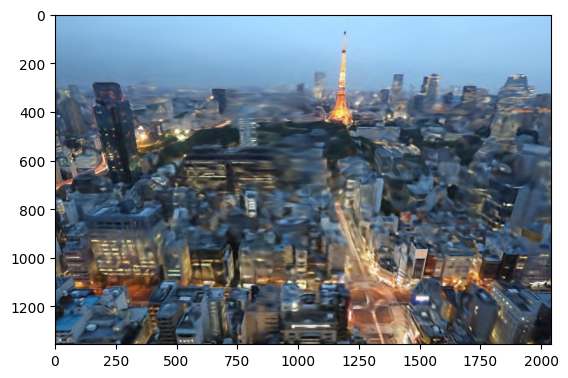}
    \caption{G 3: 20.67}
    \end{subfigure} 
    \begin{subfigure}{0.32\linewidth}  \includegraphics[width=\linewidth]{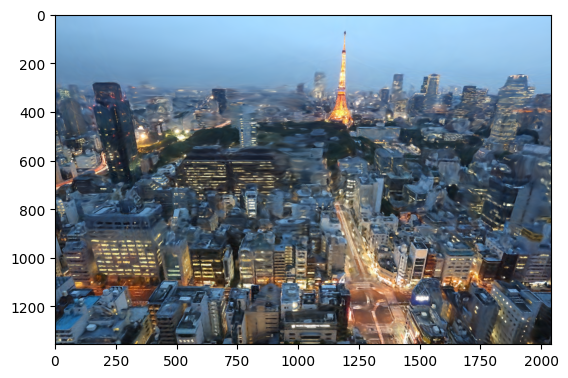}
   \caption{G 4: 21.97}
   \end{subfigure}
       \begin{subfigure}{0.32\linewidth}  \includegraphics[width=\linewidth]{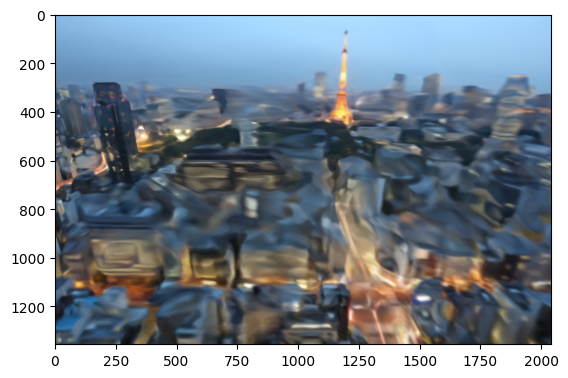}
   \caption{SGDL: 19.09}   
   \end{subfigure}
   \begin{subfigure}{0.32\linewidth}    \includegraphics[width=\linewidth]{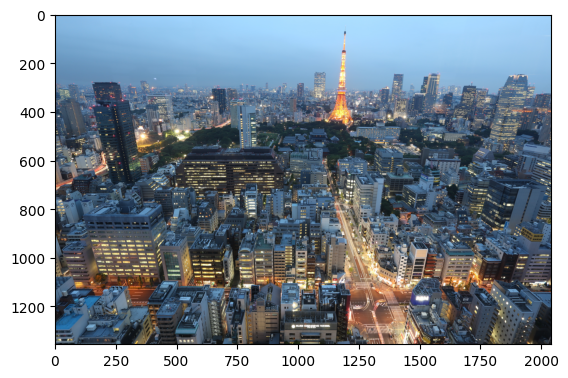}
    \caption{Ground Truth}
    \end{subfigure} 
\caption{
Comparison of MGDL and SGDL for image Building. (a)-(d): Predictions of MGDL for grades 1-4, with the corresponding testing PSNR values indicated in the subtitles. (e): Prediction of SGDL with testing PSNR displayed in the subtitle. (f): Ground truth image} 
\label{fig: 2d image regression building}
\end{figure}

The supporting figures for this experiment include  Figures \ref{fig: 2d image regression cat}-\ref{fig: 2d image regression building}, which display the predicted image for MGDL and SGDL corresponding to the Cat, Sea, and Building images, respectively.

\subsection{Section 3.4}\label{Experimental details 3.4}

The network structure for SGDL is
\begin{equation}\label{single equal structure MNIST}
[784] \to [128]\times 6  \to [10].    
\end{equation}
For MGDL, we consider two grade splittings. In the first splitting, we split network \eqref{single equal structure MNIST} into three grades, each with two hidden layers. The structure of MGDL for this splitting is as follows: 
\begin{equation}\label{MNIST-structure1}
\begin{aligned}
&\text{Grade 1:}\ \ [784] \to [128]\times 2  \to [10]\\ & \text{Grade 2:}\ \ [784] \to [128]_F\times 2 \to [128]\times 2  \to [10]\\
    &\text{Grade 3:}\ \ [784] \to [128]_F \times 4 \to  [128] \times 2 \to [10].&
\end{aligned}
\end{equation}
In the second splitting, we split network \eqref{single equal structure MNIST} into six grades, with each grade containing one hidden layer. The structure for MGDL for this case is as follows:
\begin{equation}\label{MNIST-structure2}
\begin{aligned}
&\text{Grade 1:}\ \ [784] \to [128]  \to [10]\\ & \text{Grade 2:}\ \ [784] \to [128]_F \to [128]  \to [10]\\
    &\text{Grade 3:}\ \ [784] \to [128]_F \times 2 \to  [128]\to [10]\\
    &\text{Grade 4:}\ \ [784] \to [128]_F \times 3 \to  [128]\to [10] \\
    &\text{Grade 5:}\ \ [784] \to [128]_F \times 4 \to  [128]\to [10] \\  
    &\text{Grade 6:}\ \ [784] \to [128]_F \times 5 \to  [128]\to [10] \\    
\end{aligned}
\end{equation}
For choices of parameters $t_{min}$ and $t_{max}$, we let $I_1:= \left\{10^{-4}, 10^{-5}\right\}$ and $I_2:=\left\{ 10^{-3},  10^{-4}\right\}$. For both SGDL and MGDL, we test $(t_{min}, t_{max})$ from all possible cases in the set $I_1 \times I_2$, choose the batch size from 512, 1024, or the full gradient for each epoch, and set the total number of the epochs $K$ to be 2,000.

\begin{figure}
  \centering
    \centering
   \begin{subfigure}{0.49\linewidth}\includegraphics[width=\linewidth]{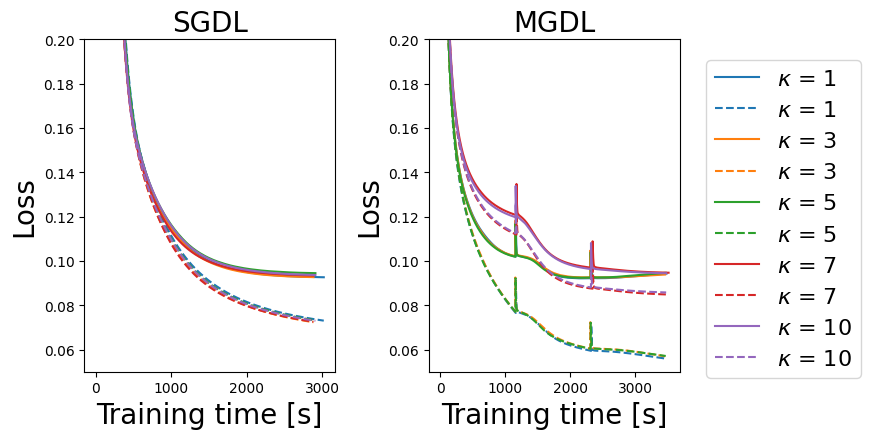}
    \caption{}
    \end{subfigure}  
   \begin{subfigure}{0.49\linewidth}  \includegraphics[width=\linewidth]{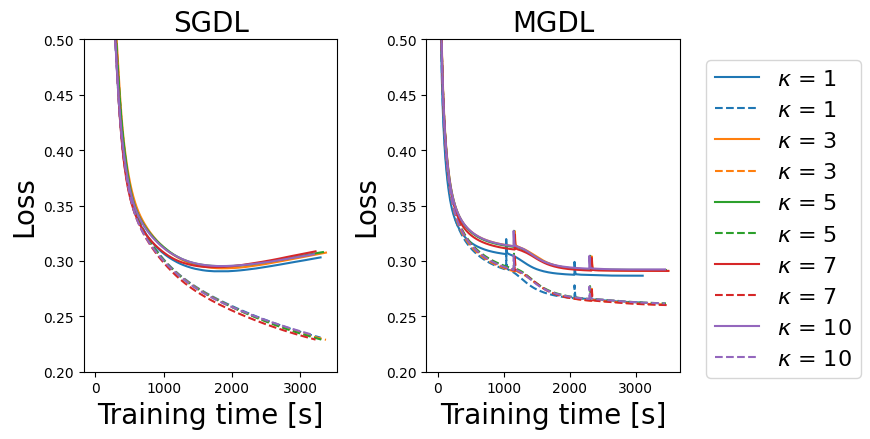}
   \caption{}
   \end{subfigure}
   \begin{subfigure}{0.49\linewidth}    \includegraphics[width=\linewidth]{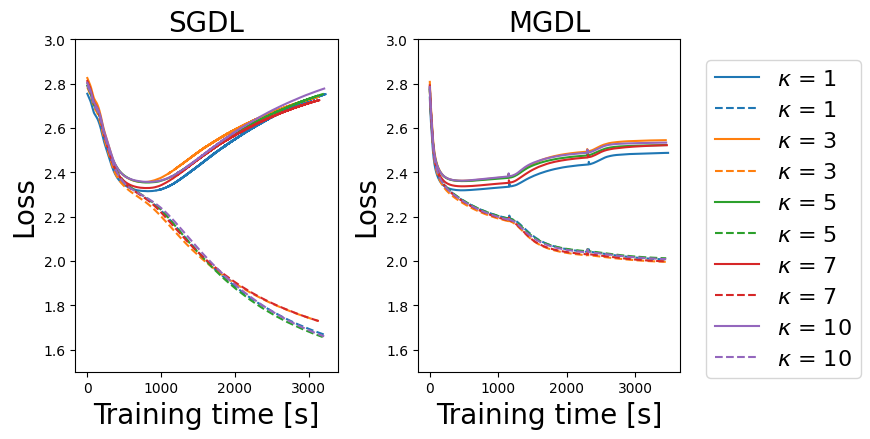}
    \caption{}
    \end{subfigure}  
   \begin{subfigure}{0.49\linewidth}  \includegraphics[width=\linewidth]{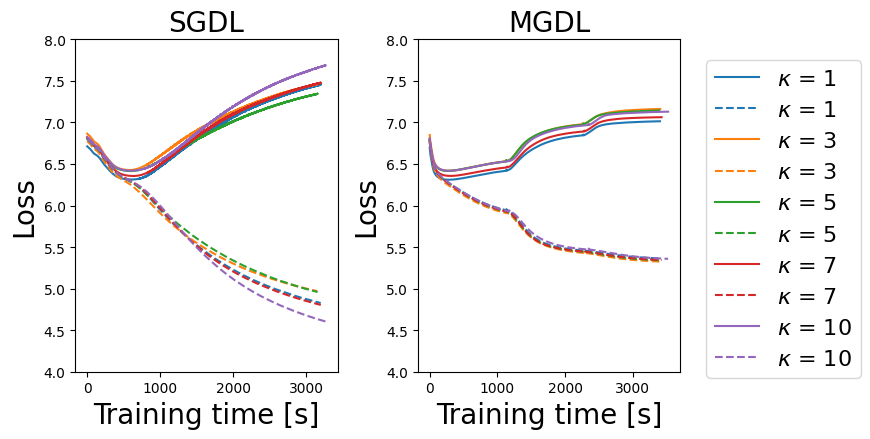}
   \caption{}
   \end{subfigure}
\caption{
Comparison of SGDL and MGDL with structure \eqref{MNIST-structure1}: training (dash curve) and validation (solid curve) loss versus training time for various values of $\beta$ and $\kappa$: (a) $\beta = 0.5$, (b) $\beta=1$, (c) $\beta = 3$, (d) $\beta = 5$.
}	\label{fig: MINIST combined SDNN MDNN vLOSS}
\end{figure}

\begin{figure}
  \centering
    \centering
   \begin{subfigure}{0.49\linewidth}\includegraphics[width=\linewidth]{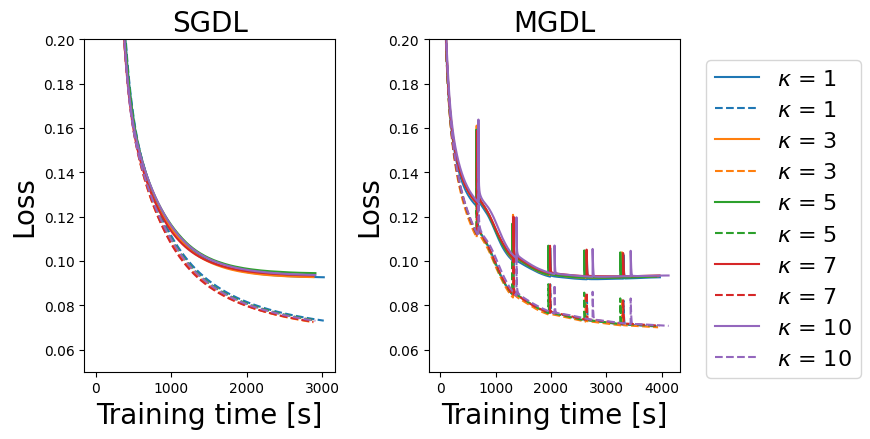}
    \caption{}
    \end{subfigure}  
   \begin{subfigure}{0.49\linewidth}  \includegraphics[width=\linewidth]{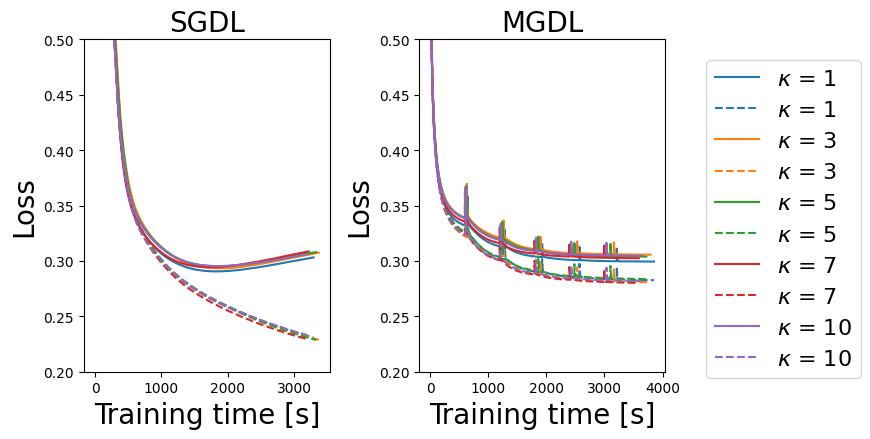}
   \caption{}
   \end{subfigure}
   \begin{subfigure}{0.49\linewidth}    \includegraphics[width=\linewidth]{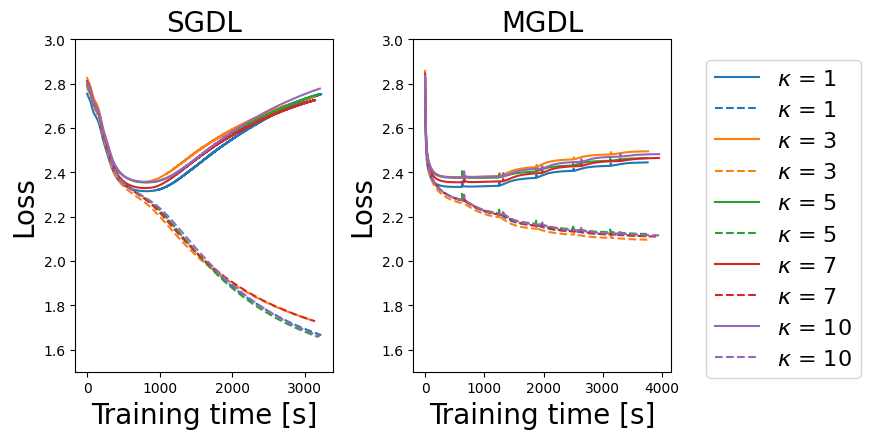}
    \caption{}
    \end{subfigure} 
       \begin{subfigure}{0.49\linewidth}  \includegraphics[width=\linewidth]{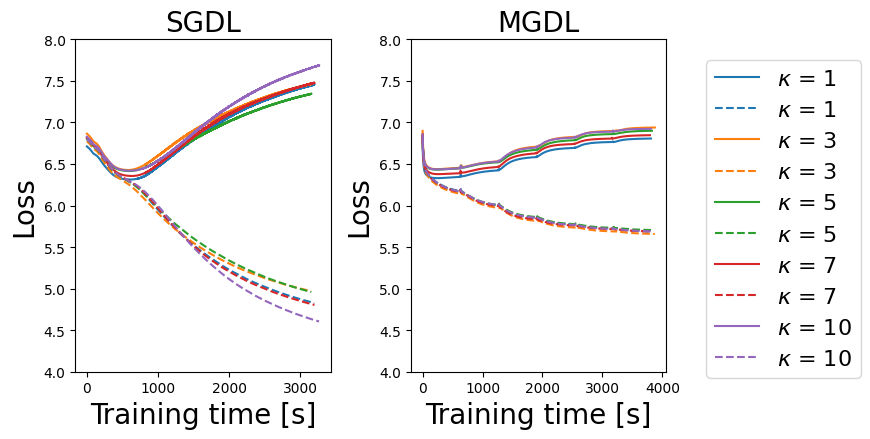}
   \caption{}
   \end{subfigure}

\caption{
Comparison of SGDL and MGDL with structure \eqref{MNIST-structure2}: training (dash curve) and validation loss (solid curve) versus training time for varies values of $\beta$ and $\kappa$: (a) $\beta = 0.5$, (b) $\beta=1$, (c) $\beta = 3$, (d) $\beta = 5$. 
}	\label{fig: MINIST combined SDNN MDNN vLOSS REDO}
\end{figure}

The supporting figures for this experiment include Figure \ref{fig: MINIST combined SDNN MDNN vLOSS}, which compares the training and validation loss for SGDL and MGDL with structure \eqref{MNIST-structure1}; Figure \ref{fig: MINIST combined SDNN MDNN vLOSS REDO}, which compares the training and validation loss for SGDL and MGDL with structure \eqref{MNIST-structure2}.

\end{document}